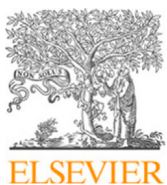

Full length article

# Modeling structured data learning with Restricted Boltzmann machines in the teacher–student setting

Robin Thériault [a],*, Francesco Tosello [b],[1], Daniele Tantari [b]

[a] *Scuola Normale Superiore di Pisa, Piazza dei Cavalieri 7, 56126, Pisa, Italy*
[b] *Department of Mathematics, University of Bologna, Piazza di Porta San Donato 5, 40126, Bologna, Italy*

A B S T R A C T

Restricted Boltzmann machines (RBM) are generative models capable to learn data with a rich underlying structure. We study the teacher–student setting where a student RBM learns structured data generated by a teacher RBM. The amount of structure in the data is controlled by adjusting the number of hidden units of the teacher and the correlations in the rows of the weights, a.k.a. patterns. In the absence of correlations, we validate the conjecture that the performance is independent of the number of teacher patterns and hidden units of the student RBMs, and we argue that the teacher–student setting can be used as a toy model for studying the lottery ticket hypothesis. Beyond this regime, we find that the critical amount of data required to learn the teacher patterns decreases with both their number and correlations. In both regimes, we find that, even with a relatively large dataset, it becomes impossible to learn the teacher patterns if the inference temperature used for regularization is kept too low. In our framework, the student can learn teacher patterns one-to-one or many-to-one, generalizing previous findings about the teacher–student setting with two hidden units to any arbitrary finite number of hidden units.

## 1. Introduction

Restricted Boltzmann machines (RBM) (Ackley, Hinton, & Sejnowski, 1985; Freund & Haussler, 1991; Hinton, 2002; Smolensky, 1986) are empirically known to fit complicated data and then sample new instances that are faithful to the underlying distribution (Hinton, 2002; Le Roux, Heess, Shotton, & Winn, 2011; Tubiana & Monasson, 2017). For example, Salakhutdinov, Mnih, and Hinton (2007) trained RBMs to predict the interests of Netflix users by fitting a database of movie ratings, Kivinen and Williams (2012) used them to generate realistic textures and Srivastava, Salakhutdinov, and Hinton (2013) employed them for topic modeling. There are even universal approximation theorems stating that an RBM with an arbitrary number of hidden units can approximate any distribution with binary support arbitrarily well (Le Roux & Bengio, 2008; Montufar & Ay, 2011). The current theoretical understanding of RBMs is largely based on statistical mechanics. For example, Decelle, Fissore, and Furtlehner (2017, 2018) used statistical mechanics to model RBM training on real data. The statistical mechanics community also studied RBMs trained on simpler, synthetic datasets (Decelle et al., 2017, 2018). In particular, many works investigated the *teacher-student setting* where a *student* RBM is trained with data produced by a *teacher* RBM (Barra, Genovese, Sollich, & Tantari, 2017, 2018; Decelle, Hwang, Rocchi, & Tantari, 2021; Hou, Wong, & Huang, 2019; Huang, 2017, 2018; Huang & Toyoizumi, 2016; Manzan & Tantari, 2024). Such studies are crucial to isolate individual characteristics of structured datasets and neural network (NN) design choices in a controlled environment and explain their effects on NN training. For example, Barra et al. (2017, 2018), Manzan and Tantari (2024) investigated the effects of the prior chosen for the data on RBM learning. In this paper, we study the effects of data correlations and number of hidden units on RBMs in the teacher–student setting. In particular, we show that RBMs with a few hidden units in the teacher–student setting can serve as a toy model of the lottery ticket hypothesis (Frankle & Carbin, 2019; Frankle, Dziugaite, Roy, & Carbin, 2020; Malach, Yehudai, Shalev-Schwartz, & Shamir, 2020; Ramanujan, Wortsman, Kembhavi, Farhadi, & Rastegari, 2020; Zhou, Lan, Liu, & Yosinski, 2019).

RBMs have a visible layer $\sigma = \{\sigma_i\}_{i=1}^{N}$, a hidden layer $\tau = \{\tau_\mu\}_{\mu=1}^{P}$ and a set of internal connections $\xi = \{\xi_i^\mu\}_{1 \leq i \leq N}^{1 \leq \mu \leq P}$, which are commonly referred to as weights. The visible layer is a set of concrete features found directly in the data, the hidden layer is an internal representation of the data in terms of abstract concepts, and the weights represent how the input features and abstract concepts are correlated with one

---

another. To give a caricatural example, the presence of pointy ears as an input feature could be correlated with the abstract concept of a cat in the weights of a particular RBM. Given $\xi$, the visible and hidden layer follow the joint distribution

$$P_\beta\left(\sigma, \tau | \xi\right) = Z_\beta(\xi)^{-1} P(\sigma) P(\tau) \exp(-\beta H[\sigma, \tau; \xi]), \quad (1)$$

where the Hamiltonian $H[\sigma, \tau; \xi] = -\frac{1}{\sqrt{N}} \sum_{\mu=1}^{P} \tau_\mu \sum_{i=1}^{N} \xi_i^\mu \sigma_i$ weighs the cost of every RBM configuration, $P(\sigma)$ and $P(\tau)$ are priors on the visible and hidden layers, respectively, and $Z_\beta(\xi) = \mathbb{E}_{\sigma,\tau}\left[\exp(-\beta H[\sigma, \tau; \xi])\right]$ is the partition function normalizing the distribution, with $\mathbb{E}_{\sigma,\tau}$ the joint expectation over the visible and hidden unit priors. Intuitively, the priors are the default distributions of $\sigma$ and $\tau$ when the Hamiltonian does not contribute to $P_\beta\left(\sigma, \tau | \xi\right)$, i.e. when $\beta$ is zero. Gibbs distributions of the form (1) have been deeply investigated in the mathematical physics community for their link with the Parisi theory of spin glasses (Barra, Contucci, Mingione, & Tantari, 2015; Barra, Genovese, Guerra, & Tantari, 2012; Genovese & Tantari, 2016, 2017, 2020; Panchenko, 2015). An RBM with a given $\xi$ can generate data $\sigma$ by sampling the marginal distribution

$$P_\beta\left(\sigma | \xi\right) = Z_\beta(\xi)^{-1} P(\sigma) \psi_\beta(\sigma; \xi) = Z_\beta(\xi)^{-1} P(\sigma) \prod_{\mu=1}^{P} \phi_\beta(\xi^\mu \cdot \sigma), \quad (2)$$

where $\phi_\beta(\xi^\mu \cdot \sigma) = \mathbb{E}_{\tau_\mu}\left[\exp\left(\frac{\beta}{\sqrt{N}} \tau_\mu \sum_{i=1}^{N} \xi_i^\mu \sigma_i\right)\right]$, $\psi_\beta(\sigma; \xi)$ factorizes as $\psi_\beta(\sigma; \xi) = \prod_{\mu=1}^{P} \phi_\beta(\xi^\mu \cdot \sigma)$ and $Z_\beta(\xi) = \mathbb{E}_\sigma\left[\psi_\beta(\sigma; \xi)\right]$. Marginal Gibbs distributions of the form (2) are also known as generalized Hopfield networks (Agliari et al., 2015a, 2015b; Agliari, Migliozzi, & Tantari, 2018; Barra et al., 2017, 2018; Hopfield, 1982; Sollich, Tantari, Annibale, & Barra, 2014). Conversely, following Bayes' theorem, the weights of an RBM can be trained on a dataset $\sigma = \{\sigma_i^a\}_{1 \leq i \leq N}^{1 \leq a \leq M}$ of $M$ examples by sampling the posterior distribution

$$P_\beta(\xi | \sigma) = \mathcal{Z}_\beta(\sigma)^{-1} P(\xi) \prod_{a=1}^{M} P_\beta(\sigma^a | \xi), \quad (3)$$

where $P(\xi)$ is a prior on the weights and $\mathcal{Z}_\beta(\sigma) = \mathbb{E}_\xi\left[\prod_{a=1}^{M} P_\beta(\sigma^a | \xi)\right]$ is the posterior partition function normalizing the distribution. In the teacher–student setting, the data $\sigma$ used to train the RBM is produced by another RBM (Barra et al., 2017, 2018; Decelle et al., 2021; Hou et al., 2019; Huang, 2017, 2018; Huang & Toyoizumi, 2016; Manzan & Tantari, 2024). In other words, a *student* RBM is trained using a dataset supplied by a *teacher* RBM. The student's ability to fit the teacher's data can then be evaluated in terms of the so-called overlaps $Q(\xi^{*\mu}, \xi^\nu) = \frac{1}{N} \sum_{i=1}^{N} \xi_i^{*\mu} \xi_i^\nu$ between the teacher's weights $\xi^* = \{\xi_i^{*\mu}\}_{1 \leq i \leq N}^{1 \leq \mu \leq P^*}$ and the student's weights $\xi = \{\xi_i^\nu\}_{1 \leq i \leq N}^{1 \leq \nu \leq P}$, whose rows $\xi^{*\mu} = \{\xi_i^{*\mu}\}_{i=1}^{N}$ and $\xi^\mu = \{\xi_i^\nu\}_{i=1}^{N}$ are also called patterns. As such, we refer to the expected value of $Q(\xi^{*\mu}, \xi^\nu)$ as the student's *performance*. This framework was used to predict the critical amount of data needed to train multiple variants of RBMs (Barra et al., 2017, 2018; Hou et al., 2019; Huang, 2017, 2018; Huang & Toyoizumi, 2016; Manzan & Tantari, 2024).

In Barra et al. (2017), it was observed empirically and conjectured that the performance of a student RBM is independent of the number of hidden units when the teacher patterns are uncorrelated. It was later shown analytically that a student RBM taught by a teacher with uncorrelated patterns achieves the same performance whether the teacher and the student have 1 or 2 hidden units each (Hou et al., 2019). In a nutshell, this simplification occurs because an RBM with 2 uncorrelated hidden units effectively factorizes into 2 RBMs with 1 hidden unit each. The conjecture of Barra et al. (2017) was explicitly rephrased in terms of this factorization property in Hou et al. (2019)[2]. Beyond this idealized scenario, it has long been known that machine learning models benefit from the correlations found in structured data to build an efficient internal representation (Gardner, 1988). This phenomenon has received considerable attention in recent theoretical studies (Chen, Min, Belkin, & Karbasi, 2021; d'Ascoli, Gabrié, Sagun, & Biroli, 2021; Decelle et al., 2017, 2018; Dobriban & Wager, 2018; Ghorbani, Mei, Misiakiewicz, & Montanari, 2021; Goldt, Mézard, Krzakala, & Zdeborová, 2020; Ichikawa & Hukushima, 2022; Liao, Couillet, & Mahoney, 2021; Loureiro et al., 2021; Nakkiran, Venkat, Kakade, & Ma, 2021; Refinetti, Goldt, Krzakala, & Zdeborova, 2021; Richards, Mourtada, & Rosasco, 2021; Wu & Xu, 2020), notably for RBMs with 2 hidden units in the teacher–student setting (Hou et al., 2019). Extending the latter study to an arbitrary number of hidden units remains an intriguing open problem. In fact, although increasing the number of hidden units in the teacher model may bring us closer to the complexity of real data—complexity that RBMs are known to capture due to universal approximation theorems (Le Roux & Bengio, 2008; Montufar & Ay, 2011)—there is still no theoretical framework that fully explains the learning performance of RBMs for an arbitrary number of hidden units.

---

[2] See the end of Appendix C of the cited paper.

In this work, we evaluate the student's learning performance in the teacher–student setting where both the teacher and the student have an arbitrary finite number of hidden units and the teacher patterns are allowed to be correlated with one another. In particular, we evaluate the critical data load $\alpha_{\text{crit}} = \frac{M}{N}$ above which learning becomes possible. In Section 2, we introduce the teacher–student setting and the replica method used in our calculations. In Section 3.1, we present the so-called *saddle-point* equations governing the performance and the critical load that we obtain from it. In Section 3.2, we discuss the case where there are no correlations between the teacher patterns. This Section is divided into three Subsections. In Sections 3.2.1 and 3.2.2, we show that the student's performance is independent of the number of hidden units when the teacher patterns are uncorrelated. In Section 3.2.3, we argue that our teacher–student setting with uncorrelated teacher patterns can serve as a toy model of the lottery ticket hypothesis. Next, in Section 3.3, we discuss the effects of uniform teacher pattern correlations on the student's performance. In particular, in Section 3.3.1, we discuss the different learning phases of the teacher–student problem as a function of the correlations and the number of hidden units. Finally, in Section 3.4, we discuss random correlations and compare their effect on the performance to that of uniform correlations. Throughout the paper, we compare key results against Monte Carlo simulations. The code and hyperparameter values of the training algorithms used to make the figures are available at the following public Github repository (Thériault, 2025).

## 2. Model

In the teacher–student setting, a student RBM with marginal likelihood $P_\beta(\sigma|\xi)$ (Eq. (2)) is trained using a dataset $\sigma = \{\sigma_i^a\}_{1 \leq i \leq N}^{1 \leq a \leq M}$ of $M$ examples $\sigma^a = \{\sigma_i^a\}_{i=1}^{N}$ of dimension $N$, generated by a teacher RBM with a prescribed marginal likelihood $P_{\beta^*}(\sigma|\xi^*)$. We call $\alpha = \frac{M}{N}$ the ratio between the size of the training set and the input dimension. In this scenario, the student knows that the correct model for the data is an RBM. However, it does not necessarily know the number of hidden units $P^*$ and the inverse temperature $\beta^*$ used by the teacher. Therefore, unless explicitly stated otherwise, we will assume that the number of hidden units $P$ and the inverse inference temperature $\beta$ of the student are not necessarily the same as the teacher's. For convenience, we will frequently state our results in terms of the temperatures $T^* = 1/\beta^*$ and $T = 1/\beta$ rather than in terms of $\beta^*$ and $\beta$.

For simplicity, we assume that the visible and hidden units of both the student and the teacher take values in $\{-1, +1\}$ with a uniform prior, i.e. they are binary random variables with no prior bias towards $-1$ or $+1$. We impose structure in the data by taking the teacher patterns to be random variables with a fixed covariance matrix $\mathcal{Q}$. To be more precise, we assume that the columns $\xi_i^* = \{\xi_i^{*\mu}\}_{\mu=1}^{P^*}$ of $\xi^*$ are i.i.d.

random variables, with mean 0 and a well-defined $P^* \times P^*$ covariance matrix $\mathcal{Q}$. Uncorrelated teacher patterns are obtained by setting $\mathcal{Q} = \mathbf{I}$.

As previously mentioned, the student learns its patterns $\xi$ by sampling them from the posterior distribution

$$
\begin{aligned}
P_\beta(\xi|\sigma) &= \mathcal{Z}_\beta(\sigma)^{-1} P(\xi) \prod_{a=1}^{M} P_\beta(\sigma^a|\xi) = \mathcal{Z}_\beta(\sigma)^{-1} P(\xi) \prod_{a=1}^{M} \left[ Z_\beta(\xi)^{-1} \psi_\beta(\sigma^a; \xi) \right] \\
&= \mathcal{Z}_\beta(\sigma)^{-1} P(\xi) \prod_{a=1}^{M} \left[ Z_\beta(\xi)^{-1} P(\sigma^a) \prod_{\mu=1}^{P} \phi_\beta(\sigma^a \cdot \xi^\mu) \right] \\
&= \mathcal{Z}_\beta(\sigma)^{-1} P(\sigma) Z_\beta(\xi)^{-M} \prod_{\mu=1}^{P} \psi_\beta(\xi^\mu; \sigma).
\end{aligned}
\quad (4)
$$

Compared with Eq. (2), the posterior distribution is composed of $P$ generalized Hopfield distributions, one for each hidden unit. The data now plays the role of dual patterns, which interact only through the term $Z_\beta(\xi)^{-M}$. The latter is therefore responsible for encoding mutual correlation between student patterns, otherwise independent (Alemanno, Camanzi, Manzan, & Tantari, 2023; Barra et al., 2017; Decelle et al., 2021). The student has no access to the structure of the teacher patterns, so we assume an identity covariance matrix for the student pattern prior $P(\xi)$.

In general, the partition functions $Z_\beta(\xi)$ and $\mathcal{Z}_\beta(\sigma)$ are intractable, which makes the properties of Eq. (4) difficult to study. However, our assumptions about the priors on the visible units, hidden units and patterns make analytical computations possible in the limit of $N, M \to \infty$.

The overlaps $Q(\xi^{*\mu}, \xi^\nu) = \frac{1}{N} \sum_{i=1}^{N} \xi_i^{*\mu} \xi_i^\nu$ are a good measure of the student's learning performance because they quantify how close the student patterns are to the teacher patterns. We exploit techniques from statistical mechanics to compute the expected value $\mathbb{E}_{\xi^*, \sigma, \xi} [Q(\xi^{*\mu}, \xi^\nu)]$ of the overlaps with respect to the distribution of the teacher patterns, the generated dataset and the inferred student patterns. Specifically, in the limit of large dataset and data dimension $M, N \to \infty$, we obtain the expected value of the overlaps as a byproduct of the limiting quenched free entropy

$$
f(\alpha, \beta^*, \beta, P^*, P, \mathcal{Q}) = \lim_{M, N \to \infty} \frac{1}{N} \mathbb{E}_{\xi^*, \sigma} \log \left[ \mathcal{Z}_\beta(\sigma) \right], \quad (5)
$$

where $\mathbb{E}_{\xi^*, \sigma}$ is the expected value w.r.t. the joint distribution of the teacher patterns and generated dataset. In fact, we show in Section 3.1 that Eq. (5) can be expressed as the result of a variational principle w.r.t. a set of order parameters, whose optimum gives the expected value of the overlaps. We then use this result to investigate the effects of $P$ and $\beta$ on the student's ability to learn a dataset characterized by $P^*, \beta^*$ and $\mathcal{Q}$, as well as the impact of these hyperparameters on the critical threshold $\alpha_{\text{crit}}$ beyond which learning becomes possible. We focus our quantitative discussion on Gaussian and binary priors $P(\xi), P(\xi^*)$ (see Appendix A.1), but our results can be generalized to other pattern distributions that meet the other assumptions stated earlier in this Section. The Gaussian case is particularly interesting because it is closely related to RBMs used in practical applications (Kivinen & Williams, 2012; Salakhutdinov et al., 2007; Srivastava et al., 2013), which usually have continuous weights.

## 3. Results and discussion

### 3.1. Free entropy and saddle point equations

The free entropy $f$ can be computed by exploiting a well-established statistical mechanics technique called the replica trick, which is based on the identity

$$
f = \lim_{N \to \infty} \frac{1}{N} \mathbb{E} \log \left[ \mathcal{Z} \right] = \lim_{N \to \infty, L \to 0} \left( \frac{1}{LN} \log \mathbb{E} \left[ \mathcal{Z}^L \right] \right).
$$

Calculations are shown in the Appendices. In Appendix B, we calculate the average replicated partition function $\mathbb{E}_{\xi^*, \sigma} \left[ \mathcal{Z}^L \right]$ in the limit of $N \to \infty$. In Appendix C, we use $\mathbb{E}_{\xi^*, \sigma} \left[ \mathcal{Z}^L \right]$ to evaluate the quenched free entropy $f$ under the so-called replica-symmetric (RS) approximation. We find that the RS free entropy can be expressed in terms of the variational principle

$$
\begin{aligned}
f = \text{Extr}_{m, \hat{m}, q, \hat{q}, s, \hat{s}} \Big\{ &- \sum_{\mu=1}^{P^*} \sum_{\nu=1}^{P} m^{\mu\nu} \hat{m}^{\mu\nu} - \frac{1}{2} \sum_{\mu \neq \nu}^{P} s^{\mu\nu} \hat{s}^{\mu\nu} + \frac{1}{2} \sum_{\mu, \nu=1}^{P} q^{\mu\nu} \hat{q}^{\mu\nu} \\
&+ \mathbb{E}_{\xi^*} \mathbb{E}_z \log \left[ \mathcal{Z} \left( \mathcal{L}^C \right) \right] + \alpha \left\langle \mathbb{E}_z \log \left[ \mathcal{Z} \left( \mathcal{L}^O \right) \right] \right\rangle_{\mathcal{M}_*} - \alpha \log \left[ \mathcal{Z}(\mathcal{M}) \right] \Big\},
\end{aligned}
\quad (6)
$$

where $z = [z_{\mu\nu}]_{\mu, \nu=1}^{P}$ is a set of i.i.d. standard Gaussian random variables, and where the effective energies $\mathcal{M}_*, \mathcal{M}, \mathcal{L}^O$ and $\mathcal{L}^C$ are defined as

$$
\mathcal{M}_*(\tau_*) = \frac{1}{2} [\beta^*]^2 \sum_{\mu, \nu=1}^{P^*} \mathcal{Q}_{\mu\nu} \tau_\mu^* \tau_\nu^* \quad (7)
$$

$$
\mathcal{M}(\tau) = \frac{1}{2} \beta^2 \sum_{\mu, \nu=1}^{P} s^{\mu\nu} \tau_\mu \tau_\nu \quad (8)
$$

$$
\mathcal{L}^O(\tau; \tau^*, z) = \mathcal{L}_{\beta^*, \beta}(\tau, \tau^*, z; m, s, q) \quad (9)
$$

$$
\mathcal{L}^C(\xi; \xi^*, z) = \mathcal{L}_{1,1}(\xi, \xi^*, z; \hat{m}, \hat{s}, \hat{q}) \quad (10)
$$

with

$$
\mathcal{L}_{\lambda_1, \lambda_2}(\xi, \xi^*, z; m, s, q) = \frac{1}{2} [\lambda_2]^2 \sum_{\mu, \nu=1}^{P} (s^{\mu\nu} - q^{\mu\nu}) \xi^\mu \xi^\nu \quad (11)
$$

$$
+ \lambda_1 \lambda_2 \sum_{\mu=1}^{P^*} \sum_{\nu=1}^{P} m^{\mu\nu} \xi^{*\mu} \xi^\nu \quad (12)
$$

$$
+ \lambda_2 \sum_{\mu, \nu=1}^{P} A_{\mu\nu}(q) z_{\mu\nu} \frac{\xi^\mu + \xi^\nu}{2}
$$

and

$$
A_{\mu\nu}(q) = \sqrt{2q^{\mu\nu} - \delta_{\mu\nu} \sum_{\eta=1}^{P} q^{\mu\eta}}. \quad (13)
$$

In particular, we use the notation $\mathcal{Z}(f) = \mathbb{E}_x \left[ \exp \{ f(x; \cdot) \} \right]$ for the partition function and $\langle g \rangle_f = \mathbb{E}_x \{ g(x) \mathcal{P}[f](x; \cdot) \}$ for the expectation value of an observable $g$ w.r.t. the Gibbs distribution $\mathcal{P}[f](x; \cdot) = \mathcal{Z}(f)^{-1} \exp \left[ f(x; \cdot) \right]$ of an effective energy $f$, where $\mathbb{E}_x$ is the expectation over the prior of $x$. As explained in Appendix A.3, $q^{\mu\nu}$ is assumed symmetric in Eq. (11). The solution of this variational principle (see Appendix D for a detailed derivation) must obey the saddle-point equations

$$
\begin{aligned}
m^{\mu\nu} &= \mathbb{E}_{\xi^*} \mathbb{E}_z \left[ \xi^{*\mu} \langle \xi^\nu \rangle_{\mathcal{L}^C} \right] \\
s^{\mu\nu} &= \mathbb{E}_{\xi^*} \mathbb{E}_z \left[ \langle \xi^\mu \xi^\nu \rangle_{\mathcal{L}^C} \right] \\
q^{\mu\nu} &= \mathbb{E}_{\xi^*} \mathbb{E}_z \left[ \langle \xi^\mu \rangle_{\mathcal{L}^C} \langle \xi^\nu \rangle_{\mathcal{L}^C} \right], \\
\hat{m}^{\mu\nu} &= \beta^* \beta \alpha \left\langle \mathbb{E}_z \left[ \tau_\mu^* \langle \tau_\nu \rangle_{\mathcal{L}^O} \right] \right\rangle_{\mathcal{M}_*} \\
\hat{s}^{\mu\nu} &= \beta^2 \alpha \left( \left\langle \mathbb{E}_z \left[ \langle \tau_\mu \tau_\nu \rangle_{\mathcal{L}^O} \right] \right\rangle_{\mathcal{M}_*} - \left\langle \tau_\mu \tau_\nu \right\rangle_{\mathcal{M}} \right) \\
\hat{q}^{\mu\nu} &= \beta^2 \alpha \left\langle \mathbb{E}_z \left[ \langle \tau_\mu \rangle_{\mathcal{L}^O} \langle \tau_\nu \rangle_{\mathcal{L}^O} \right] \right\rangle_{\mathcal{M}_*}.
\end{aligned}
\quad (14)
$$

The optimal order parameters $m \in \mathbb{R}^{P^* \times P}$, $s \in \mathbb{R}^{P \times P}$ and $q \in \mathbb{R}^{P \times P}$ solving Eqs. (14) are to be interpreted as expected overlaps. First of all, $m^{\mu\nu}$, which is commonly known as the Mattis magnetization, is the limiting expected overlap between the teacher pattern $\xi^{*\mu}$ and the student pattern $\xi^\nu$, i.e.

$$
m^{\mu\nu} = \lim_{N, M \to \infty} \mathbb{E}_{\xi^*, \sigma, \xi} \left[ Q(\xi^{*\mu}, \xi^\nu) \right] \quad \mu = 1, \ldots, P^* \; \nu = 1, \ldots, P, \quad (15)
$$

where $\mathbb{E}_{\xi^*, \sigma, \xi}$ is the expectation w.r.t. the joint distribution $P(\xi^*) P_{\beta^*}(\sigma|\xi^*) P_\beta(\xi|\sigma)$ of teacher patterns, dataset and student patterns. Second of all, $s^{\mu\nu}$ with $\mu \neq \nu$ is the limiting expected overlap between any two student patterns $\xi^\mu$ and $\xi^\nu$ from the same sample $\xi \sim P_\beta(\xi|\sigma)$, i.e.

$$
s^{\mu\nu} = \lim_{N, M \to \infty} \mathbb{E}_{\xi^*, \sigma, \xi} \left[ Q(\xi^\mu, \xi^\nu) \right] \quad \mu, \nu = 1, \ldots, P. \quad (16)
$$

Finally, $q^{\mu\nu}$ is the limiting expected overlap between any two student patterns $\xi^{1\mu}$ and $\xi^{2\nu}$ from two independent posterior samples $\xi^1$ and $\xi^2$, i.e.

$$q^{\mu\nu} = \lim_{N,M \to \infty} \mathbb{E}_{\xi^*, \sigma, \xi^1 \times \xi^2} \left[ Q(\xi^{1\mu}, \xi^{2\nu}) \right]$$
$$= \lim_{N,M \to \infty} \mathbb{E}_{\xi^*, \sigma} \left[ Q(\mathbb{E}_{\xi|\sigma} \left[ \xi^\mu \right], \mathbb{E}_{\xi|\sigma} \left[ \xi^\nu \right]) \right] \quad \mu, \nu = 1, \ldots, P, \quad (17)$$

where $\mathbb{E}_{\xi|\sigma}$ indicates the expectation w.r.t. the posterior distribution (Eq. (4)). While $s$ measures the effective correlation between student patterns, the so-called spin-glass order parameter $q$ quantifies the tendency of the student to stay frozen in specific configurations rather than sampling all possible patterns uniformly. For simplicity's sake, we will usually call $m$ the magnetization and $q$ the spin-glass (SG) overlap. As in the Introduction, we will also occasionally refer to $m$ as the student's performance.

The RS saddle point equations (Eq. (14)) can be solved by numerical iteration for any values of the hyperparameters $\beta^*$, $\beta$, $\alpha$, $P^*$ and $P$ (see Appendix I). We expect the RS solution to be exact when the student is fully informed about the teacher's prior and hyperparameters and matches them with its own, i.e. $\beta = \beta^*$, $P = P^*$ and $P(\xi) = P(\xi^*)$. This regime of complete information is commonly referred to as the Nishimori line (Contucci, Giardinà, & Nishimori, 2009; Iba, 1999; Nishimori, 1980, 2001). When $\beta = \beta^*$, we find two different phases (see Figs. 2, 3, 10 and 16):

- the *paramagnetic* (P) phase, where the order parameters $m$, $s$ and $q$ all vanish;
- the *ferromagnetic* (F) phase, where they are all larger than zero.

Intuitively, the student RBM can partially learn the teacher patterns in the F phase but becomes unable to do so in the P phase. The corresponding P-F phase transition is thus also the onset of learning. When $\beta \neq \beta^*$, we also find

- the *spin-glass* (SG) phase, where $m = 0$ but $q > 0$.

In this phase, the student converges to spurious low-energy states ($q > 0$) unrelated to the teacher patterns ($m = 0$) (see Figs. 11 and 12). Looking at Figs. 2, 3, 10 and 16, the P-F phase transition appears to be second order. In general, such second-order phase transitions coincide with the onset of instability of the paramagnetic solution in the saddle-point equations (Eqs. (14)). The instability condition (see Appendix F) reads as

$$\alpha \geq \alpha_{\text{crit}} = \frac{1}{[\beta^* \beta]^2 \lambda_{\max}^S}, \quad (18)$$

where $\lambda_{\max}^S$ is the largest eigenvalue of the matrix $S = Q\mathcal{R}$, with the covariance matrix $\mathcal{R}$ defined as $\mathcal{R}_{\mu\nu} = \langle \tau^{*\mu} \tau^{*\nu} \rangle_{\mathcal{M}_*}$. As expected, the $\alpha_{\text{crit}}$ of Eq. (18) coincides with the P-F phase transition of Figs. 10 and 16. Interestingly, $\alpha_{\text{crit}}$ does not depend on the number of hidden units $P$ of the student nor on its prior $P(\xi)$. Outside the Nishimori line, i.e. when $(\beta, P, P(\xi)) \neq (\beta^*, P^*, P(\xi^*))$, the RS approximation is not expected to be exact. Therefore, in principle, one would need to calculate the so-called replica symmetry breaking (RSB) corrections of $\alpha_{\text{crit}}$. However, we find that Eq. (18) is consistent with Monte Carlo simulations even when $P \neq P^*$ (see Fig. 5) and $P(\xi) \neq P(\xi^*)$ (see Fig. 14). As such, the RS approximation of $\alpha_{\text{crit}}$ seems robust to the priors and number of hidden units in practice.

### 3.2. Learning uncorrelated patterns

In this Section, including Sections 3.2.1–3.2.3, we take $Q = \mathbf{I}$, i.e. there are no correlations between the teacher patterns. In that regime, we find that the teacher–student setting exhibits an effective factorization property on the student patterns, which makes the behavior of the student RBM explainable in terms of that with a single hidden unit. A first indication of this result is given by the critical load $\alpha_{\text{crit}}$ (see Eq. (18)). As we explained in the previous Section, no matter $Q$,

Eq. (18) does not depend on the number $P$ of student patterns. When $Q = \mathbf{I}$, and thus $S = \mathbf{I}$ (see Eqs. (18)), we find that

$$\alpha_{\text{crit}} = \frac{1}{[\beta^* \beta]^2}, \quad (19)$$

which does not depend on the number $P^*$ of teacher patterns either. Eq. (19) generalizes to arbitrary finite $P^*$ and $P$ the critical load $\alpha_{\text{crit}} = \beta^{-4}$ previously found for $P^* = P = 1$ hidden units (Huang, 2017; Huang & Toyoizumi, 2016) and $P^* = P = 2$ uncorrelated hidden units (Hou et al., 2019) when $\beta^* = \beta$. As we show in the next Subsections, the previous universality result also holds for the order parameters, which we recall are a good measure of the student's learning performance.

#### 3.2.1. Independence of the number of hidden units: binary patterns

In the case of binary patterns, Eqs. (14) with $P = P^* = 1$ simplify to

$$\begin{aligned}
m &= \mathbb{E}_z \left[ \tanh\left(\hat{m} + \sqrt{\hat{q}} z\right) \right] \\
q &= \mathbb{E}_z \left[ \tanh^2\left(\hat{m} + \sqrt{\hat{q}} z\right) \right] \\
\hat{m} &= \beta^* \beta \alpha \, \mathbb{E}_z \left[ \tanh\left(\beta^* \beta m + \beta \sqrt{q} z\right) \right] \\
\hat{q} &= \beta^2 \alpha \, \mathbb{E}_z \left[ \tanh^2\left(\beta^2 m + \beta \sqrt{q} z\right) \right],
\end{aligned} \quad (20)$$

whose solution is summarized in the phase diagrams of Fig. 1. To be more precise, the left and right panels of Fig. 1 show the fixed-$\beta^*$ regime and the Nishimori regime $\beta = \beta^*$, respectively. For clarity's sake, we will discuss Fig. 1 and the solution of Eq. (20) in terms of $T^* = 1/\beta^*$ and $T = 1/\beta$ for the remainder of this paragraph. We recall that the student can learn the teacher patterns in the ferromagnetic (F) phase ($m \neq 0$), but not in the paramagnetic (P) phase ($m = q = 0$) nor in the spins-glass (SG) phase ($m = 0$ and $q > 0$). For $T < T^*$, we find the SG phase between the P phase and the F phase. The corresponding P-SG transition line $\alpha_{\text{P-SG}} = T^4$ extends until the Nishimori line $T = T^*$, where it meets the F phase. Therefore, the Nishimori line crosses the triple point of the P, SG and F phases, as expected from spin glass theory (Nishimori, 1980, 2001). On the Nishimori line, it is straightforward to verify that $m = q$, which is also expected from spin glass theory (Nishimori, 1980, 2001). In particular, $m$ and $q$ simultaneously become non-zero above the critical load $\alpha_{\text{crit}} = [T^*T]^2 = T^4$ of the P-F phase transition on the Nishimori line, which prevents the SG phase from forming (see Fig. 1, right panel). As in some related inference problems in the teacher–student setting, $\alpha_{\text{P-SG}} = T^4$ is identical to the $\alpha_{\text{crit}}$ found on the Nishimori line (Alemanno et al., 2023; Manzan & Tantari, 2024; Thériault & Tantari, 2024). For $T > T^*$, the SG phase does not exist either, and the P phase transitions directly to the F phase at $\alpha_{\text{crit}} = [T^*T]^2$.

Looking at the SG-F transition, we see that decreasing the inference temperature $T$ too much can make it harder for the student to learn the teacher patterns. In particular, a student with enough data to learn the teacher patterns at a given inference temperature $T_1$ may fail to do so at a lower inference temperature $T_2 < T_1$. The RS approximation is probably not completely accurate in this regime, so one would need to calculate RSB corrections to study this behavior quantitatively.

When $P, P^*$ are arbitrary finite numbers and $Q = \mathbf{I}$, we search for solutions of Eqs. (14) by making the ansatz

$$\begin{aligned}
m^{\mu\nu} &= \delta_{\mu\nu} m, & \hat{m}^{\mu\nu} &= \delta_{\mu\nu} \hat{m}, \\
s^{\mu\nu} &= \delta_{\mu\nu}, & \hat{s}^{\mu\nu} &= 0, \\
q^{\mu\nu} &= \begin{cases} \delta_{\mu\nu} q & \mu, \nu \leq P^* \\ \delta_{\mu\nu} g & \text{otherwise,} \end{cases} & \hat{q}^{\mu\nu} &= \begin{cases} \delta_{\mu\nu} \hat{q} & \mu, \nu \leq P^* \\ \delta_{\mu\nu} \hat{g} & \text{otherwise,} \end{cases}
\end{aligned} \quad (21)$$

which describes the situation where the student learns the teacher patterns one-to-one. Following the nomenclature introduced in Hou et al. (2019), we will call Eq. (21) the permutation symmetry breaking (PSB) ansatz. We have assumed without loss of generality that the $\min\{P, P^*\}$ non-zero magnetizations are on the main diagonal, as any hidden unit permutation would give an equivalent solution. According to this ansatz, the first $\min\{P, P^*\}$ student patterns converge one-to-one to the first $\min\{P, P^*\}$ teacher patterns with magnetization $m$ and SG overlap $q$. When $P > P^*$, the remaining $P - P^*$ student patterns (i.e. $P \geq \nu > P^*$) are aligned in spurious directions (i.e. $m^{\mu\nu} = 0 \ \forall \mu = 1, \ldots, P^*$) with a different SG overlap $g \neq q$. Conversely, when $P < P^*$, $P^* - P$ of the teacher patterns are not learned for lack of student hidden units. The latter case is described in terms of only $m$ and $q$, and so is $P = P^*$. Under the PSB ansatz, Eqs. (14) decouple into $P$ independent systems of equations for the $P$ hidden units $\nu = 1, \ldots, P$ (see Appendix G). In other words, the student factorizes into $P$ students with one hidden unit each. The systems of equations for the first $\min\{P, P^*\}$ patterns are all identical to each other and equivalent to Eqs. (20). When $P > P^*$, the $P - P^*$ remaining systems of equations all take the form

$$g = \mathbb{E}_z\left[\tanh^2\left(\sqrt{\hat{g}}z\right)\right] \tag{22}$$
$$\hat{g} = \beta^2 \alpha \ \mathbb{E}_z\left[\tanh^2\left(\beta\sqrt{g}z\right)\right].$$

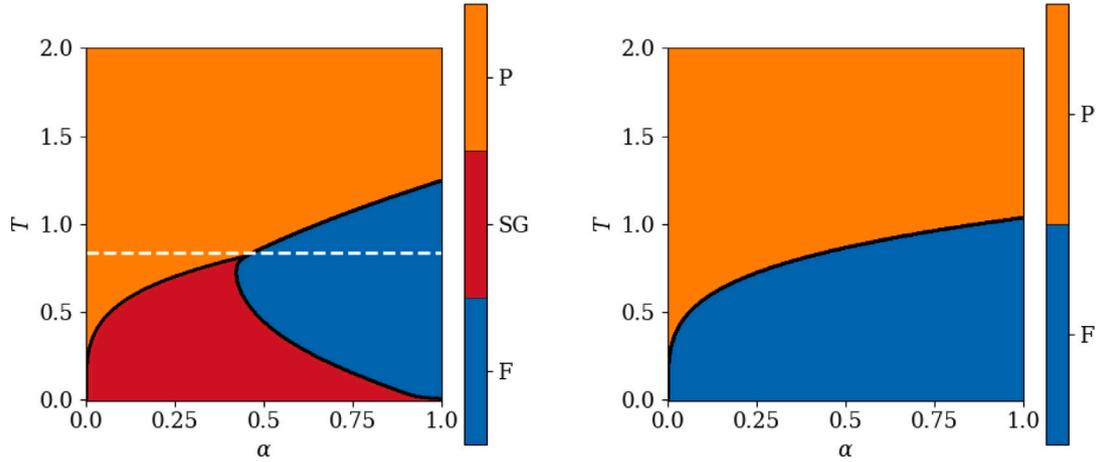

**Fig. 1.** RS phase diagrams of the teacher–student setting with $P = P^* = 1$ obtained by solving Eqs. (20). The left diagrams has $\beta^* = 1.2$, and the right one, $\beta^* = \beta = 1/T$. The student is unable to learn the teacher patterns in the paramagnetic phase (P) and in the spin-glass phase (SG), but it is able to do so in the ferromagnetic phase (F). The white dashed line in the left plot is the Nishimori line $\beta^* = \beta$. The magnetization $m$ and SG overlap $q$ solving Eqs. (20) along this line are plotted on Fig. 2.

In summary, Eqs. (14) with binary patterns and $Q = I$ reduce to Eqs. (20), (22) under the PSB ansatz. In particular, the solution of Eqs. (20), (22), which we will call the PSB solution, has $m = q$ even in the region outside the Nishimori line where $\beta = \beta^*$ and $P \neq P^*$. In Fig. 2, we verify that the PSB solution can be found by iterating the original saddle-point equations (Eqs. (14)) with binary student patterns (see Appendix A.1), $Q = I$ and near-diagonal initial conditions $m^{\mu\nu} = \delta_{\mu\nu}m_0 + (1 - \delta_{\mu\nu})\varepsilon$, where $\varepsilon \ll m_0$. In other words, we verify that the solution of Eqs. ((20), (22)) is stable for the full saddle-point iteration. We plot only the case of $P \geq P^*$ because $P < P^*$ yields similar results.

The original saddle-point equations (Eq. (14)) with binary patterns, $Q = I$ and $P > P^*$ also have other solutions that cannot be expressed as cleanly by a set of simplified saddle-point equations (see Fig. 3). These solutions can be found by initializing $m^{\mu\nu}$ with $m_0 \gg \varepsilon$ on the diagonal and at least one off-diagonal entry close to $m_0$, a process we will refer to as off-diagonal initialization. They correspond to a distributed representation where some of the student patterns learn the same teacher pattern, so we dub these solutions partial PSB. Throughout this paper and its figures, we focus on the case where at most two student patterns learn the same teacher pattern, but larger numbers are also possible. Within a partial PSB solution, some student patterns may still learn teacher patterns one-to-one. These student patterns $\xi_{PSB}^\mu$ have the same $m$ and $q$ as Eq. (20), and in particular satisfy $m = q$ (see Fig. 3). Comparison of the RS free entropy (Eq. 3.1) of the PSB and partial PSB solutions of Eqs. (14) suggests that the former is always favored in the limit $N, M \to \infty$. The free entropy difference between them decreases with increasing $T = T^*$ and vanishes at the onset of the paramagnetic phase (see Fig. J.17). These results are confirmed by numerical simulations. In low $T = T^*$ simulations, students with binary patterns always converge to the PSB solution, even when we use random initial conditions (see Figs. 5), which is solid evidence that PSB is favored at low $T = T^*$. When $T = T^*$ is relatively close to the paramagnetic phase, the simulations have relatively large error bars and repeatedly jump between the different solutions rather than converging to a single mode. This kind of instability can be attributed to finite-size fluctuations.

The PSB solution is independent of $P^*$ and $P$, which means that, in terms of the student's performance $m$, learning $P^*$ i.i.d. patterns stored in a single teacher RBM is as difficult as learning $P^*$ i.i.d. patterns from $P^*$ separate RBMs with one hidden unit each. Such independence of the student's performance from the number of hidden units was first conjectured in Barra et al. (2017) based on empirical observations. The performance on the Nishimori line was then shown to be the same for $P = P^* = 2$ (Hou et al., 2019) as for $P = P^* = 1$ (Huang, 2017; Huang & Toyoizumi, 2016). Our PSB solution extends this result to $\beta = \beta^*$ and any finite $P, P^*$. On the Nishimori line, i.e. when $\beta = \beta^*$ and $P = P^*$, the teacher–student setting is replica symmetric, so we expect the PSB solution to be exact, which is confirmed by simulations (see Fig. 4). When $\beta = \beta^*$ and $P \neq P^*$, replica symmetry is not guaranteed, but the PSB solution is still in good agreement with simulations (see Fig. 5). Finally, we expect RSB corrections when $\beta \neq \beta^*$. Interestingly, we observe a weaker form of independence from the number of hidden units in the suboptimal partial PSB solutions, in the sense that the partial PSB solutions that we find at given $P^*$ and $P$ are also present at larger $P^*$ and $P$. These results seem related to the embedding principle stating that a neural network contains all minima of narrower neural networks with the same architecture (Zhang, Zhang, Luo, & Xu, 2021).

#### 3.2.2. Independence of the number of hidden units: Gaussian patterns

When the student patterns $\xi$ are real-valued variables with an i.i.d. standard Gaussian prior, Eqs. (14) under the PSB ansatz (Eqs. (21))

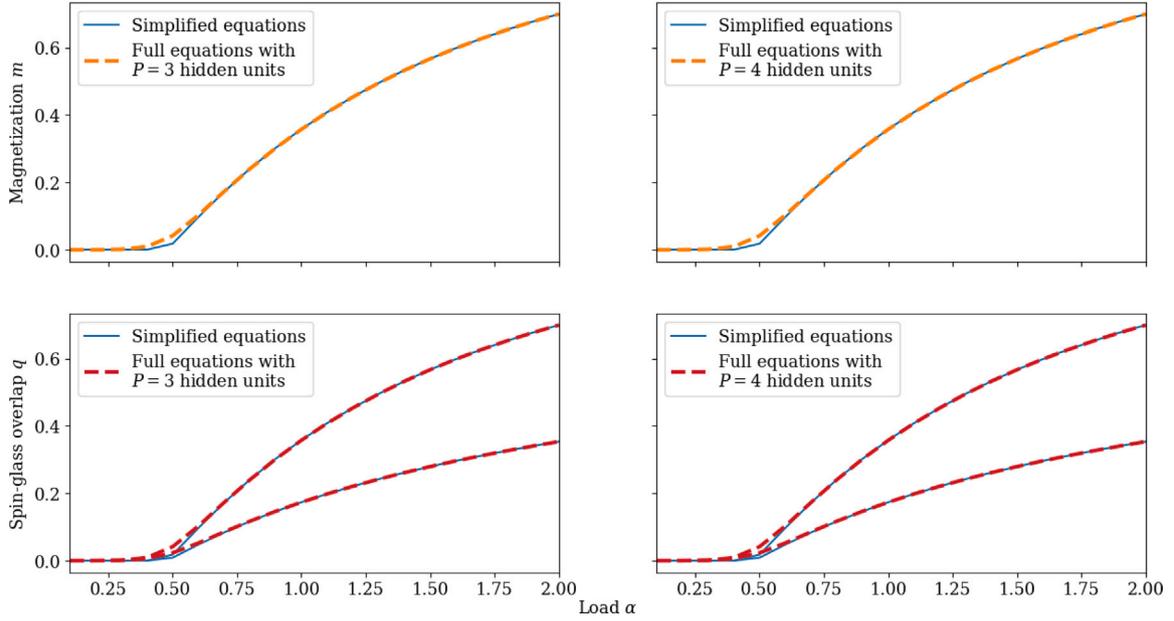

**Fig. 2.** Permutation symmetry breaking (PSB) solution of Eqs. (14) for binary student patterns with a uniform prior and binary teacher patterns with covariance $Q = \mathbf{I}$, in red and orange, compared against the solution of Eqs. ((20), (22)), in blue. We plot the Mattis magnetization $m$ in the top row, and the spin-glass (SG) overlap $q$ in the bottom row. The magnetization plots and the top lines of the SG overlap plots show that the student patterns that converge to teacher patterns have the same $m$ and $q$ as the solution of Eqs. (20), and thus also satisfy $m = q$. Conversely, the bottom lines of the SG overlap plots show that the student patterns that do not converge to a teacher pattern have the same SG overlap $g$ as the solution of Eqs. (22). We use $P = 3$ and $P^* = 2$ in the left column and $P = 4$ and $P^* = 3$ in the right column. All plots have $\beta^* = \beta = 1.2$.

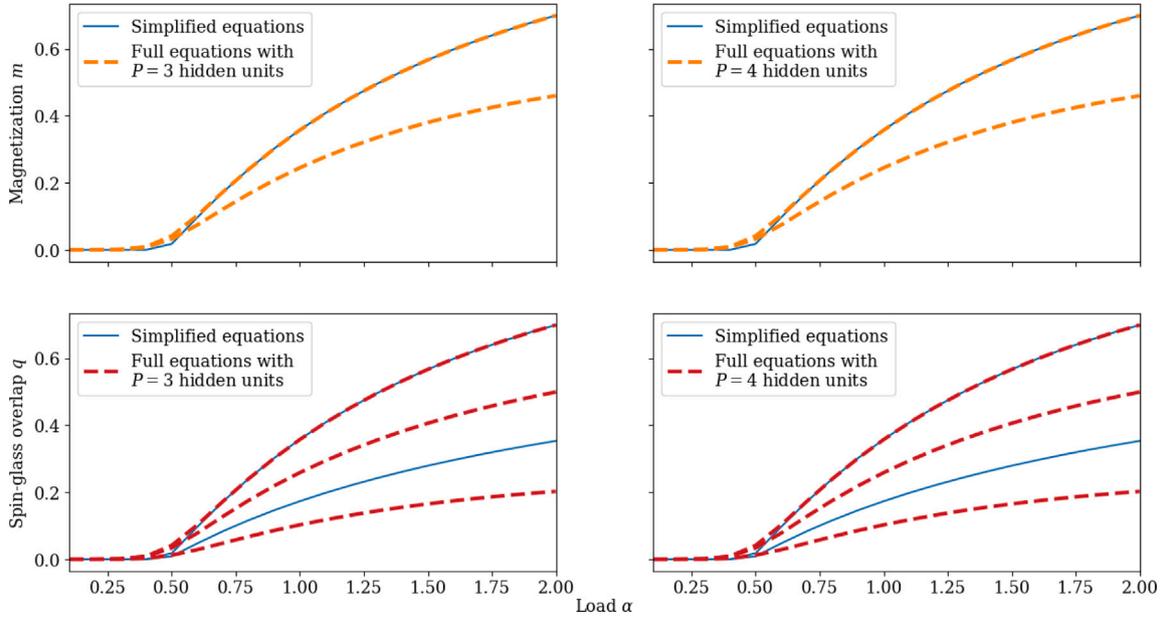

**Fig. 3.** Partial permutation symmetry breaking (partial PSB) solutions of Eqs. (14) for binary student patterns with a uniform prior and binary teacher patterns with covariance $Q = \mathbf{I}$, in red and orange, compared against the solution of Eqs. ((20), (22)), in blue. We plot the Mattis magnetization $m$ in the top row, and the spin-glass (SG) overlap $q$ in the bottom row. The top lines of the plots show that the student patterns $\xi_{\text{PSB}}^\mu$ that converge to teacher patterns one-to-one have the same $m$ and $q$ as the solution of Eqs. (20), and thus also satisfy $m = q$. Conversely, the other lines show that the student patterns $\xi_{\text{PS}}^\mu$ that converge to a common teacher pattern have a smaller $m$ and a different $q$. To be more precise, the central and bottom branches of $q$ are the spin-glass order parameters corresponding to $Q(\xi_{P,S}^{1\mu}, \xi_{P,S}^{2\mu})$ and $Q(\xi_{P,S}^{1\mu}, \xi_{P,S}^{2\nu})$ with $\mu \neq \nu$, respectively (see Section 2). They are both different from the $g$ of Eq. (22). The Mattis magnetization and SG overlaps omitted from this Figure all vanish. We use $P = 3$ and $P^* = 2$ in the left column and $P = 4$ and $P^* = 3$ in the right column. All plots have $\beta^* = \beta = 1.2$.

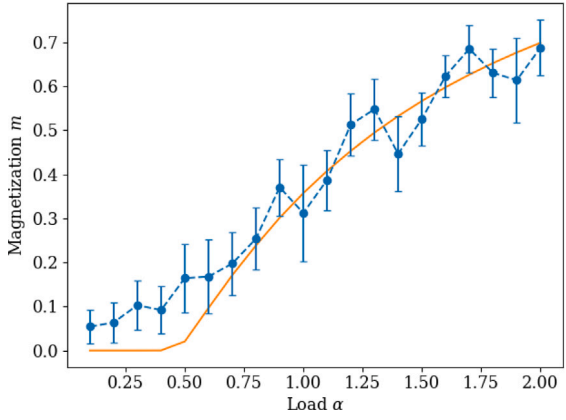

**Fig. 4.** The magnetization $m$ solving Eqs. (22), in orange, compared against $N = 512$ dimensional Monte Carlo simulations, in blue, of the teacher–student problem where the student has $P = 2$ binary patterns with a uniform prior and the teacher has $P^* = P = 2$ binary patterns with covariance $Q = \mathbf{I}$. The blue dots and error bars represent the means and standard deviations, respectively, of the diagonal of the magnetization $m$ during the simulations. The inverse temperature is set to $\beta^* = \beta = 1.2$, and the simulations have a small external field biasing the student towards the PSB solution.

simplify to (see Appendix G)

$$
\begin{aligned}
m &= \frac{\hat{m}}{1 + \hat{q}} \\
q &= \frac{\hat{m}^2}{(1 + \hat{q})^2} + \frac{\hat{q}}{(1 + \hat{q})^2} \\
g &= \frac{\hat{g}}{(1 + \hat{g})^2} \\
\hat{m} &= \beta^* \beta \alpha \; \mathbb{E}_z \left[ \tanh \left( \beta^* \beta m + \beta \sqrt{q} z \right) \right] \\
\hat{q} &= \beta^2 \alpha \; \mathbb{E}_z \left[ \tanh^2 \left( \beta^2 m + \beta \sqrt{q} z \right) \right] \\
\hat{g} &= \beta^2 \alpha \; \mathbb{E}_z \left[ \tanh^2 \left( \beta \sqrt{g} z \right) \right].
\end{aligned}
\tag{23}
$$

This regime is interesting because RBMs used in practical applications usually have continuous weights (Kivinen & Williams, 2012; Salakhutdinov et al., 2007; Srivastava et al., 2013). Similarly as before, we compare Eqs. (23) against the PSB and partial PSB solutions of Eqs. (14) for real-valued student patterns with an i.i.d. standard Gaussian prior and teacher pattern covariance $Q = \mathbf{I}$. The resulting plots are qualitatively similar to Figs. 2 and 3, so we show them in Appendix J rather than in the main text (see Figs. J.18 and J.19). As before, we observe that the solutions of the Gaussian equations are independent of $P^*$ and $P$. One important difference between the student model with binary patterns and the student model with real-valued patterns is that they are simulated differently. In fact, binary patterns are obtained using the standard random walk Metropolis–Hastings algorithm, while real-valued patterns are learned via underdamped stochastic Langevin dynamics (Welling & Teh, 2011; Zhang, Chewi, Li, Balasubramanian, & Erdogdu, 2023) of the RBM marginal likelihood (Eq. (2)) gradient. In a nutshell, the latter algorithm is a variant of stochastic gradient ascent where noise is added to the momentum vector to sample the RBM posterior (Eq. (3)) rather than to find one of its modes. We use contrastive divergence, i.e. alternate Gibbs sampling of the visible and hidden units, to evaluate the marginal likelihood (Hinton, 2002, 2012). At low temperature, simulations of the student model with real-valued patterns usually converge to the solution of Eqs. (23) (see Fig. 6). They can also sometimes stay stuck near the partial PSB solution without converging to it when the learning rate is reduced too aggressively during training. This behavior is shown in Fig. 6 by the simulation data points that are scattered around the partial PSB solution but do not agree with it within their error bars. In fact, although simulation error is useful for measuring the convergence of the Langevin learning algorithm, its sensitivity to the learning rate schedule makes it poorly representative of the equilibrium distribution of the magnetization $m$.

### 3.2.3. A simple model of the lottery ticket hypothesis

Many works studying feedforward neural networks found a training regime where some of the hidden units learn the underlying data distribution while the others take a back seat (Chizat & Bach, 2018; Luo, Xu, Ma, & Zhang, 2021; Ma, Wu, & E, 2020; Zhang, Zhang, Luo, & Xu, 2021). Interestingly, this behavior is similar to our PSB solution with $P > P^*$, where only a subset of the student patterns learn the teacher patterns (see Sections 3.2.1 and 3.2.2). As proposed in Zhang, Zhang, Luo, and Xu (2021), we investigate the relationship between this kind of training regime and the lottery ticket hypothesis (Frankle & Carbin, 2019).

The lottery ticket hypothesis states that a generic randomly initialized overparameterized neural network contains subnetworks that fit data with similar accuracy as the entire trained network when they are extracted from it and trained independently (Frankle & Carbin, 2019; Frankle et al., 2020; Malach et al., 2020; Ramanujan et al., 2020; Zhou et al., 2019). These special subnetworks, which are commonly referred to as winning lottery tickets, are thought to have fortuitous initial conditions that facilitate training (Frankle & Carbin, 2019). The PSB solution studied in Sections 3.2.1 and 3.2.2 makes it clear that any student network of size $P > P^*$ contains subnetworks of size $\tilde{P} \in \{P^*, \ldots, P - 1\}$ that learn the teacher patterns to the same extent, and thus fit the data at least as well. It is not obvious, however, whether any of these subnetworks can have lucky initial conditions such that they train more easily than with i.i.d. random initial conditions. As such, we apply to our teacher–student setting a variant of the magnitude pruning algorithm traditionally used to identify winning tickets (Frankle & Carbin, 2019; Frankle et al., 2020) and check if it finds a subnetwork that converges especially quickly. Consider one teacher and three students with real-valued patterns, hereby referred to as *the teacher*, *student 0*, *student A* and *student B*, respectively, where A is a control network and B is a candidate winning ticket obtained from 0. We perform the following numerical experiment:

- Initialize student 0 with $P = 8$ i.i.d. Gaussian patterns and the teacher with $P^* = 4$ i.i.d. Gaussian patterns. Save the initial value of the patterns of student 0 as $\xi_0$.
- Train student 0 on data generated by the teacher. Compute the Euclidean norms of the learned patterns.
- Initialize student A with $P = 4$ i.i.d. Gaussian patterns and student B with the $P = 4$ patterns of $\xi_0$ that evolved to have the largest Euclidean norm when trained.
- Train student A and student B on data generated by the teacher. Record their respective magnetizations $m_A$ and $m_B$ as a function of the training epochs in order to compare their convergence speeds.

This procedure is slightly different from the usual magnitude pruning algorithm in that it prunes the patterns with the smallest norms rather than the entries of the weight matrix with the smallest absolute values (Frankle & Carbin, 2019; Frankle et al., 2020). As in Section 3.2.2, we train students 0, A and B with underdamped stochastic Langevin dynamics (Welling & Teh, 2011; Zhang et al., 2023) of the RBM marginal likelihood (Eq. (2)). Our results are shown in Fig. 7. In the left panel, we verify that student B converges to the PSB solution (i.e. the solution of Eqs. (23)) in the range of $\alpha$ that we are studying. As in Section 3.2.2, the error in $m$ is useful for measuring the convergence of the learning algorithm, but poorly representative of the equilibrium distribution. There seems to be a small systematic shift in the simulations at low $\alpha$, which could be due to finite-size effects or order $\frac{P}{M}$ corrections (Barra et al., 2017). In the right panel, we plot the median of $m_B - m_A$ over $\alpha$ as a function of the epochs. B initially converges more quickly than A, and the lead of B eventually shrinks to zero because of the ergodicity of the training algorithm. In other words, magnitude pruning is able to identify a winning ticket that converges more quickly than networks with i.i.d. random initial conditions. The patterns of student 0 that converge to the teacher patterns the fastest have a small initial magnetization (see Fig. 7, epoch 0). Therefore, the shape of the loss function at the initial conditions and the basin of attraction in which they are located may be more important than the initial conditions themselves, as suggested by Frankle and Carbin (2019).

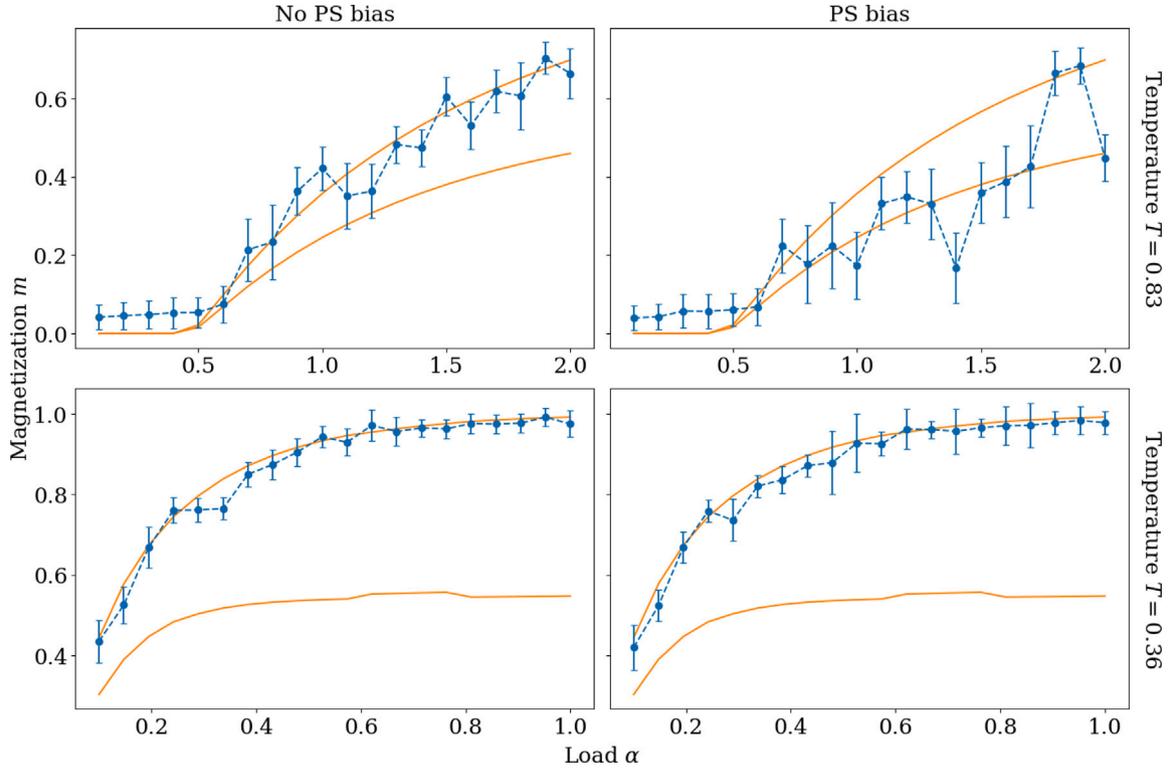

**Fig. 5.** Solutions of Eqs. (14) shown in Fig. 3, in orange, compared against $N = 512$ dimensional Monte Carlo simulations, in blue, of the teacher–student problem where the student has $P = 2$ binary patterns with a uniform prior and the teacher has a single binary pattern. The blue dots and error bars represent the means and standard deviations, respectively, of a single diagonal coefficient of $m$ during the simulations. The inverse temperature is set to $\beta^* = \beta = 1.2 \approx 1/0.83$ and $\beta^* = \beta = 2.8 \approx 1/0.36$ in the top and bottom rows, respectively. In the left column, the student with $T^* = T = 0.83$ is biased towards the PSB solution by near-diagonal initial conditions, while the student with $T^* = T = 0.36$ has random initial conditions. The students of the right column have off-diagonal initial conditions. The top left and top right students are also biased by a small external field pointing towards the PSB and partial PSB solutions, respectively. The bottom students are not biased by an external field.

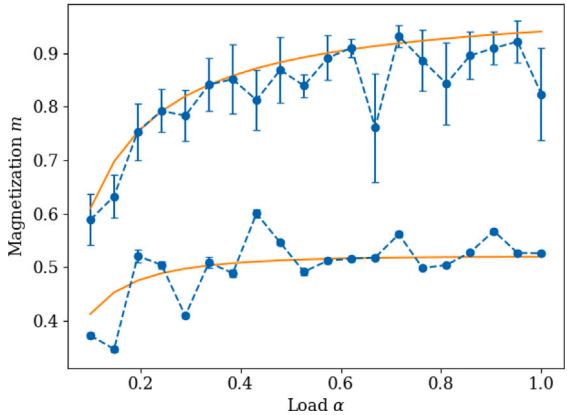

**Fig. 6.** Solutions of Eqs. (14) for real-valued student patterns with a standard Gaussian prior and teacher pattern covariance $\mathcal{Q} = \mathbf{I}$, in orange, compared against $N = 512$ dimensional Monte Carlo simulations, in blue, of the teacher–student problem where the student has $P = 2$ real-valued patterns with a standard Gaussian prior and the teacher has a single Gaussian pattern. The top orange line is the magnetization $m$ of student patterns that converge to teacher patterns one-to-one, so it is also the $m$ solving Eqs. (23). On the other hand, the bottom orange line is the $m$ of student patterns that converge to a common teacher pattern in the partial PSB solution. The blue dots and error bars represent the means and standard deviations, respectively, of the largest entry of the magnetization $m$ during the simulations. The top and bottom blue lines result from simulations with i.i.d. standard Gaussian initial conditions and off-diagonal initial conditions, respectively. The learning rate was reduced more quickly for the bottom line than for the top line. The inverse temperature is set to $\beta^* = \beta = 4$.

### 3.3. Learning uniformly correlated patterns

In this Section, including Section 3.3.1, we introduce uniform correlations in the teacher patterns by fixing the covariance matrix $\mathcal{Q}$ of $\xi^*$ to $\mathcal{Q}_{\mu\nu} = \delta_{\mu\nu} + (1 - \delta_{\mu\nu}) c$, where $c \in [0, 1)$ controls the correlation strength. In presence of uniform correlation, the Hamiltonian $\mathcal{M}_*$ (Eq. (7)) is that of the Curie–Weiss model with coupling constant $\frac{1}{2}[\beta^*]^2 c$. Therefore, its correlation matrix $\mathcal{R}$ has the same form as $\mathcal{Q}$ but with a different off-diagonal element $d$. In Appendix H, we show that the maximum eigenvalue $\lambda^S_{\max}$ of Eq. (18) is

$$\lambda^S_{\max} = (P^* - 1)^2 cd + (P^* - 1)(c + d) + 1. \quad (24)$$

As expected, the corresponding critical load is also the onset of non-zero Mattis magnetization in the regime of $\beta = \beta^*$ where only the paramagnetic and ferromagnetic phases exist, which we show explicitly in Fig. 10 for binary patterns. We show in Appendix J that the same holds for real-valued student patterns with a standard Gaussian prior (see Fig. J.20). Eq. (24) extends to arbitrary finite $P^*$ the critical load obtained for $P^* = 2$ in Hou et al. (2019). In fact, when $P^* = 2$, we find $d = \tanh\left([\beta^*]^2 c\right)$ and Eq. (18) reduces to the critical load of Hou et al. (2019). Plotting Eq. (24) (see Fig. 9) and the corresponding $\alpha_{\text{crit}}$ (Fig. 8) as a function of $T$, $c$ and $P^*$, we see that $\alpha_{\text{crit}}$ decreases with $P^*$ and $c$. For $c \gg [T^*]^2$, the spins of Curie–Weiss Hamiltonian $\mathcal{M}_*$ all align, so their correlation is $d = 1$, and we obtain

$$\lambda^S_{\max} = ((P^* - 1) c + 1) P^*.$$

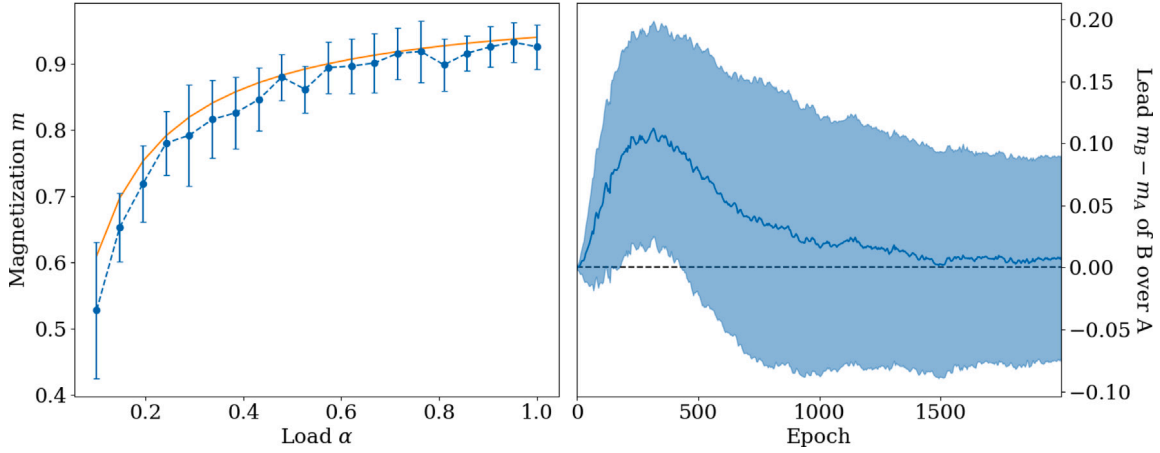

**Fig. 7.** Results of the lottery ticket experiment described in Section 3.2.3. In the left panel, $N = 512$ dimensional Monte Carlo simulations of student B, in blue, are compared against the solution of Eqs. (23), in orange. The blue dots and error bars represent the means and standard deviations, respectively, of the diagonal of the magnetization $m$ during the simulations. The right panel shows the difference $m_B - m_A$ of the magnetizations of A and B as a function of the simulation epochs. The solid blue line and the shaded region represent the median of $m_A - m_B$ over $\alpha \in [0, 1]$ and the corresponding mean absolute deviation around the median, respectively. $m_A - m_B$ goes to zero when the number of elapsed epochs is large, so student A converges to the solution of Eqs. (23) like student B. The inverse temperature is set to $\beta^* = \beta = 4$.

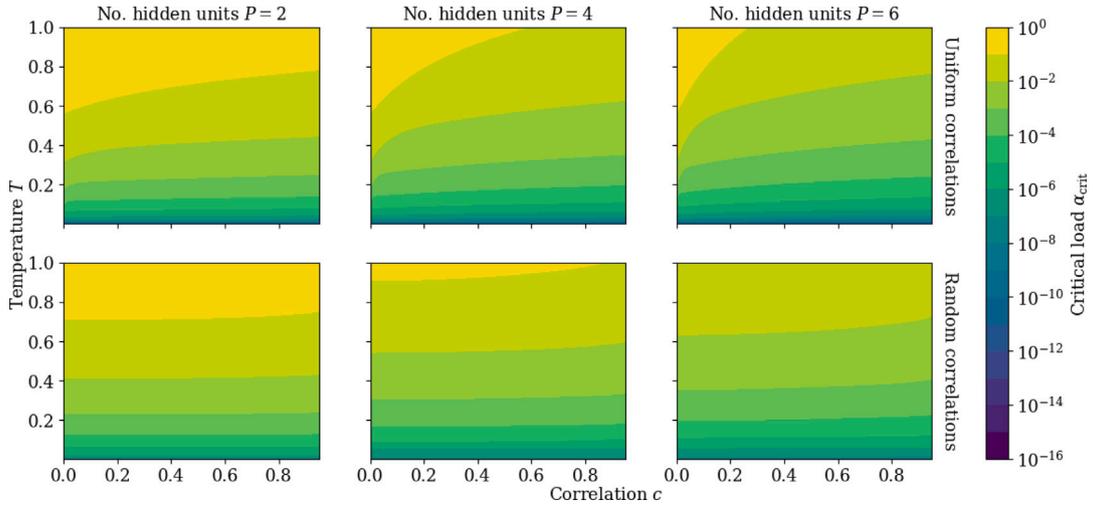

**Fig. 8.** Critical load $\alpha_{\text{crit}}$ for $\beta = \beta^*$ and $P = P^*$ as a function of the number of hidden units $P$, the temperature $T$ and the correlation $c$. $\alpha_{\text{crit}}$ is obtained from Eq. (18). The top row has $Q_{\mu\nu} = \delta_{\mu\nu} + (1 - \delta_{\mu\nu}) c$, so the max eigenvalue $\lambda_{\max}^S$ is that of Eq. (24). The bottom row is the arithmetic mean $\overline{\alpha_{\text{crit}}}$ over correlation matrices $\mathcal{Q}$ sampled from the projected Wishart distribution $\mathcal{W}(c, P)$ defined in Appendix A.2.

In this limit, $\alpha_{\text{crit}}$ is roughly inversely proportional to both $c$ and $[P^*]^2$. Conversely, for $c \ll [T^*]^2$, the correlation of $\mathcal{M}_*$ is $d = 0$, and we obtain

$$\lambda_{\max}^S = (P^* - 1) c + 1. \qquad (25)$$

In this limit, $\alpha_{\text{crit}}$ decreases less quickly by a factor of $P^*$ than for $c \gg [T^*]^2$. However, relatively small correlations can still significantly decrease $\alpha_{\text{crit}}$ when $P^*$ is large. In other words, the critical load benefits from structured data with many correlated underlying abstract concepts even when the correlation between the different concepts is rather small. Overall, these results shed light upon the way $\alpha_{\text{crit}}$ depends on $P^*$, which was previously unclear given that previous work (Hou et al., 2019) focused on $P = P^* = 2$.

Neural networks are often regularized by feature decorrelation techniques that reduce or eliminate correlations in their inputs and hidden layers (Huang, Huangi, Yang, Lang, & Deng, 2018; Ioffe & Szegedy, 2015; LeCun, Bottou, Orr, & Müller, 2012; Zhang, Nezhadarya, et al., 2021). Decorrelation is known to increase training speed (Huangi et al., 2018; Ioffe & Szegedy, 2015; Wiesler & Ney, 2011; Zhang, Nezhadarya, et al., 2021). However, our work suggests that it can also increase the critical load as a drawback, which may hinder performance when the data is noisy (i.e. $\beta^*$ finite). Fagbohungbe and Qian (2022), Galloway, Golubeva, Tanay, Moussa, and Taylor (2019) observed that batch normalization (Ioffe & Szegedy, 2015), which was motivated by decorrelation (Huangi et al., 2018; Lecun, Bottou, Bengio, & Haffner, 1998), makes neural networks less robust to noise. This effect could potentially be related to our findings. Arguably, the main caveat to our analysis is that we assumed that $P^*$ was finite when deriving Eq. (24), so the critical load is probably different when $P^*$ is $\mathcal{O}(N)$ (Barra et al., 2017).

Figs. 11 and 12 display the magnetization $m$ and the SG overlap $q$, respectively, found by solving Eqs. (14) at fixed $\beta^*$. As usual, we find the paramagnetic (P) phase ($m = q = 0$) at small $\alpha$ and low $T$. At low $c$ and $P$, the spin glass (SG) phase ($m = 0$ and $q > 0$) occupies the medium $\alpha$ and low $T$ region of the $\alpha, T$ plane. It transitions to the ferromagnetic (F) phase ($m \neq 0$) as $\alpha$, $c$ and $P$ increase. At high $c$, $P$ and $\alpha$, the SG overlap $q$ seems to be non-monotonic in $T$. As expected, the critical load of Eq. (18) with $\lambda^S_{\max}$ given by Eq. (24) follows the onset of non-zero magnetization corresponding to the P-F phase transition, but not the SG-F phase transition. The critical load of $\beta = \beta^*$ approximately follows the P-SG phase transition when $c$ is small, which is consistent with Section 3.2.1 and previous works (Alemanno et al., 2023; Thériault & Tantari, 2024). As in Section 3.2.1, decreasing the inference temperature $T$ too much can make it harder for the student to learn the teacher patterns (see Fig. 11, top right panel).

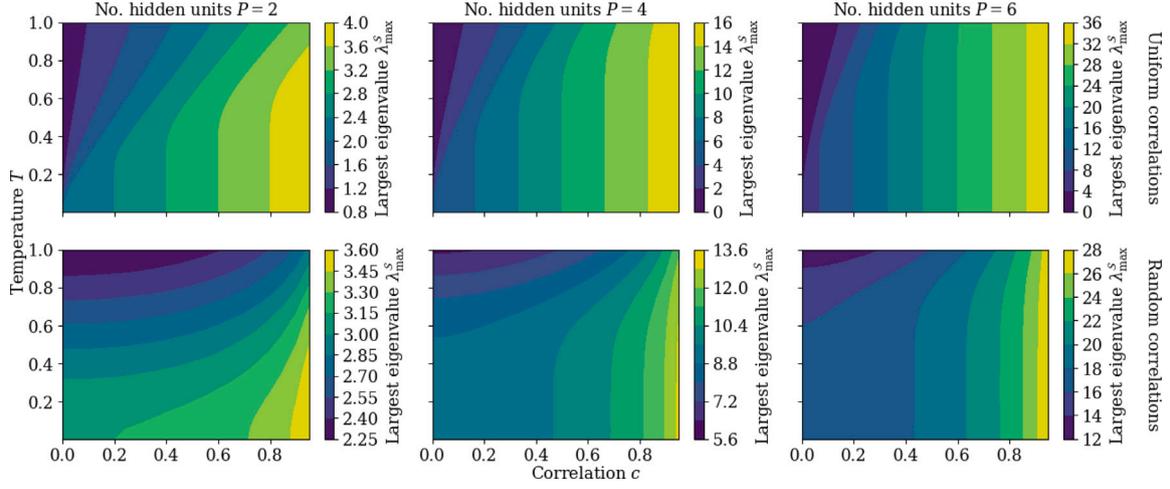

**Fig. 9.** Largest eigenvalue $\lambda^S_{\max}$ of $S = QR$ (see Appendix F) for $\beta = \beta^*$ and $P = P^*$ as a function of the number of hidden units $P$, the temperature $T$ and the correlation $c$. The top row has $Q_{\mu\nu} = \delta_{\mu\nu} + (1 - \delta_{\mu\nu})c$, so the max eigenvalue $\lambda^S_{\max}$ is that of Eq. (24). The bottom row is the harmonic mean $\left[\overline{1/\lambda^S_{\max}}\right]^{-1}$ over correlation matrices $Q$ sampled from the projected Wishart distribution $W(c, P)$ defined in Appendix A.2.

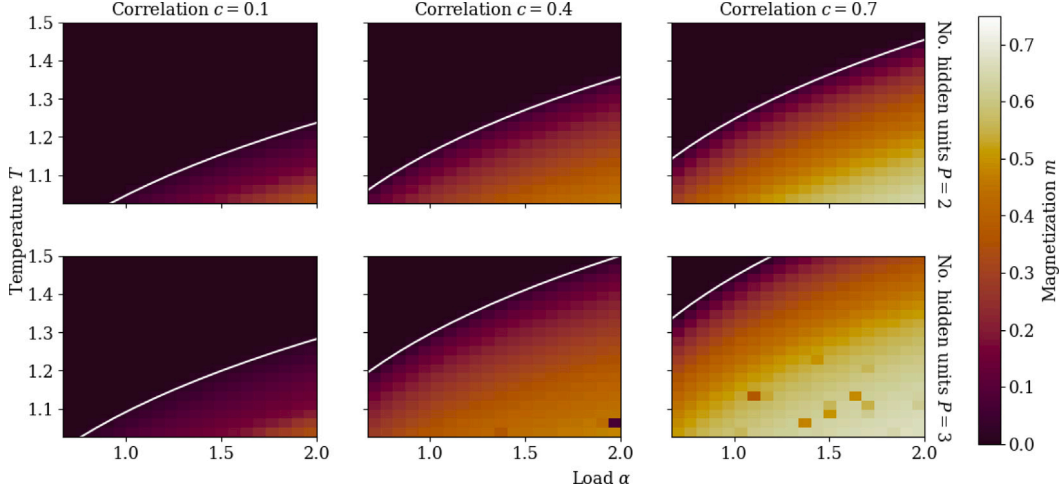

**Fig. 10.** Mattis magnetization $m$ for $\beta = \beta^*$ and $P = P^*$ as a function of the number of hidden units $P$, the correlation $c$, the temperature $T$ and the data load $\alpha$. $m$ is obtained by solving Eqs. (14) numerically for binary student patterns with a uniform prior and binary teacher patterns with covariance $Q_{\mu\nu} = \delta_{\mu\nu} + (1 - \delta_{\mu\nu})c$, where $c \in [0, 1)$ (see Appendix H). The top and bottom rows feature $P = 2$ and $P = 3$, respectively. The white lines mark the phase transition described by Eq. (18) with $\lambda^S_{\max}$ given by Eq. (24). The speckles in the plots with $P = 3$, $c = 0.4$ and $P = 3$, $c = 0.7$ are due to the occasional numerical instability of the saddle-point equations (see Appendix I).

#### 3.3.1. Permutation symmetry breaking transitions

As in Section 3.2, the critical load marking the onset of learning (see Eq. (18)) again does not depend on the number of hidden units $P$ of the student RBM. Despite this, a single wide RBM does not learn teacher patterns as would multiple separate RBMs with one hidden unit each. Moreover, distinct phases emerge based on the level of correlation $c$ and data load $\alpha$, each characterized by a different learning strategy.

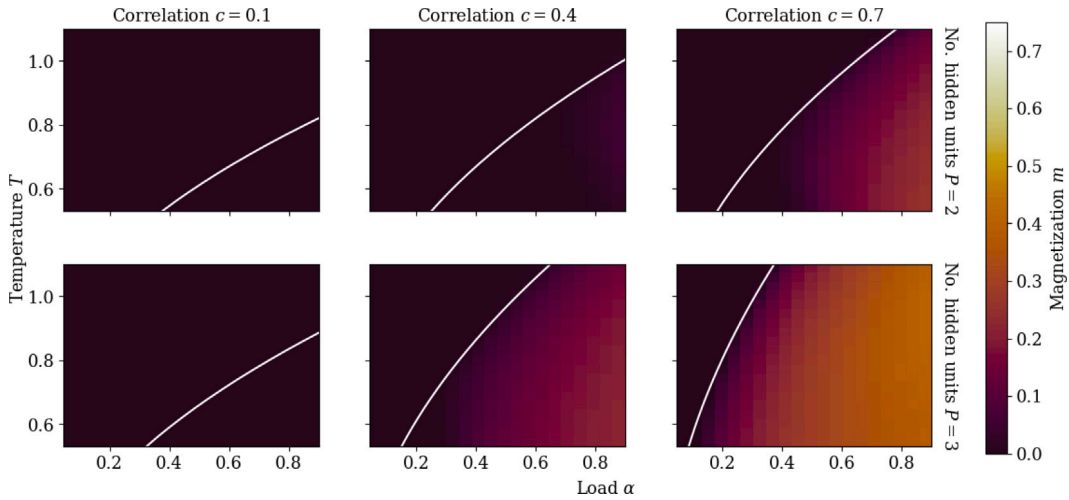

**Fig. 11.** Mattis magnetization $m$ for $\beta^* = 0.8$ and $P = P^*$ as a function of the number of hidden units $P$, the correlation $c$, the temperature $T$ and the data load $\alpha$. $m$ is obtained by solving Eqs. (14) numerically for binary student patterns with a uniform prior and binary teacher patterns with covariance $Q_{\mu\nu} = \delta_{\mu\nu} + (1 - \delta_{\mu\nu})c$, where $c \in [0,1)$ (see Appendix H). The top and bottom rows feature $P = 2$ and $P = 3$, respectively. The white lines mark the critical load of Eq. (18) with $\beta^* = 0.8$ and $\lambda_{\max}^S$ given by Eq. (24).

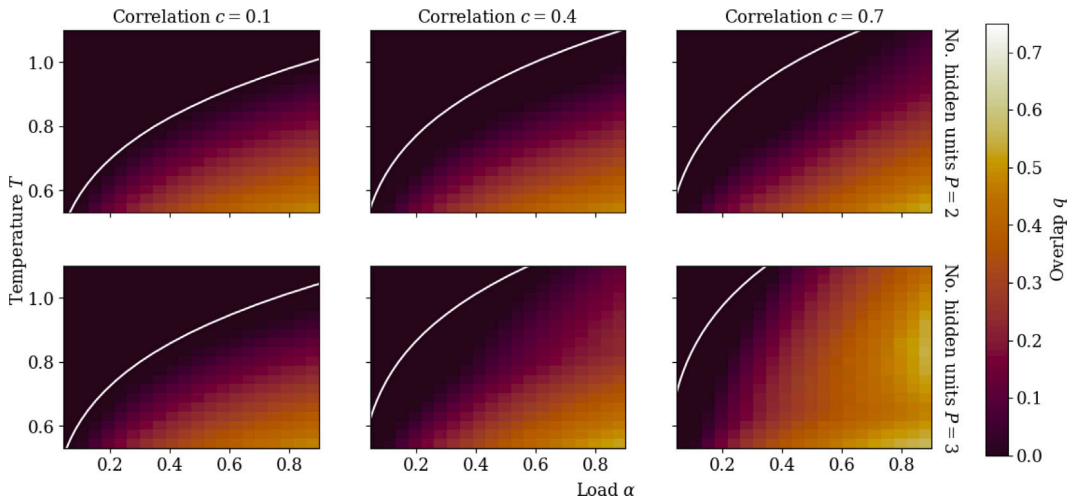

**Fig. 12.** SG overlap $q$ for $\beta^* = 0.8$ and $P = P^*$ as a function of the number of hidden units $P$, the correlation $c$, the temperature $T$ and the data load $\alpha$. $q$ is obtained by solving Eqs. (14) numerically for binary student patterns with a uniform prior and binary teacher patterns with covariance $Q_{\mu\nu} = \delta_{\mu\nu} + (1 - \delta_{\mu\nu})c$, where $c \in [0,1)$ (see Appendix H). The top and bottom rows feature $P = 2$ and $P = 3$, respectively. The white lines mark the critical load of Eq. (18) with $\beta = \beta^*$ and $\lambda_{\max}^S$ given by Eq. (24).

RBMs with $P = P^* = 2$ hidden units learning correlated patterns were previously found to have three distinct ferromagnetic phases (Hou et al., 2019):

- in the *spontaneous symmetry breaking* (SSB) phase, both student patterns converge to a single configuration that has the same overlap with the two teacher patterns. In other words, the student learns only the features that the teacher patterns have in common;
- in the *student permutation symmetry breaking* (PSB$_s$) phase, the student patterns converge to two distinct configurations that have the same overlap with the teacher patterns. In other words, the student is able to learn distinct features of the teacher patterns;
- finally, in the *teacher permutation symmetry breaking* (PSB$_t$) phase, the student can tell the two teacher patterns apart and learns them one-to-one as in the PSB phase of Section 3.2.1.

These three phases can be identified by comparing the values of the order parameters $m^{\mu\nu}$ and $q^{\mu\nu}$ on and off the diagonal. As shown in the left panel of Fig. 13, bifurcations occur as $\alpha$ increases, marking second-order phase transitions at the critical loads $\alpha_{\text{crit}}^{\text{SSB}} = \alpha_{\text{crit}}$, $\alpha_{\text{crit}}^{\text{PSB}_s} > \alpha_{\text{crit}}^{\text{SSB}}$ and $\alpha_{\text{crit}}^{\text{PSB}_t} > \alpha_{\text{crit}}^{\text{PSB}_s}$. In this context, the student does not know the level of correlation $c$ of the teacher patterns. As such, it uses a uniform prior $P(\xi) \neq P(\xi^*)$, and we are always outside the Nishimori regime. Despite this, the RS critical thresholds and magnetizations are in good agreement with Monte Carlo simulations (see Fig. 13, left panel).

As shown in Figs. 13 and 14, the same three phases also appear at larger $P$ and $P^*$. $\alpha_{\text{crit}}^{\text{SSB}}$ once again coincides with the onset of the ferromagnetic phase, which occurs at the same load $\alpha_{\text{crit}}$ regardless of the number of hidden units $P$ of the student. On the other hand, $\alpha_{\text{crit}}^{\text{PSB}_s}$ is smaller for $P = 3$ than for $P = 2$ (see Fig. 13). We think that the term $A_{\mu\mu}(q) = \sqrt{2q^{\mu\mu} - \sum_{\eta=1}^{P} q^{\mu\eta}}$ in $\mathcal{L}_{\lambda_1 \lambda_2}$ (see Eqs. (11) and (13)) penalizes values of $q$ with a large off-diagonal sum, i.e. $\sum_{\eta \neq \mu}^P q^{\mu\eta}$, compared to the diagonal $q^{\mu\mu}$. As the number of terms in the sum grows with $P$, the individual off-diagonal coefficients $q^{\mu\nu}$ may be encouraged to become smaller, pushing permutation symmetry breaking of the student patterns (PSB$_s$) to occur at a lower critical threshold $\alpha_{\text{crit}}^{\text{PSB}_s}$.

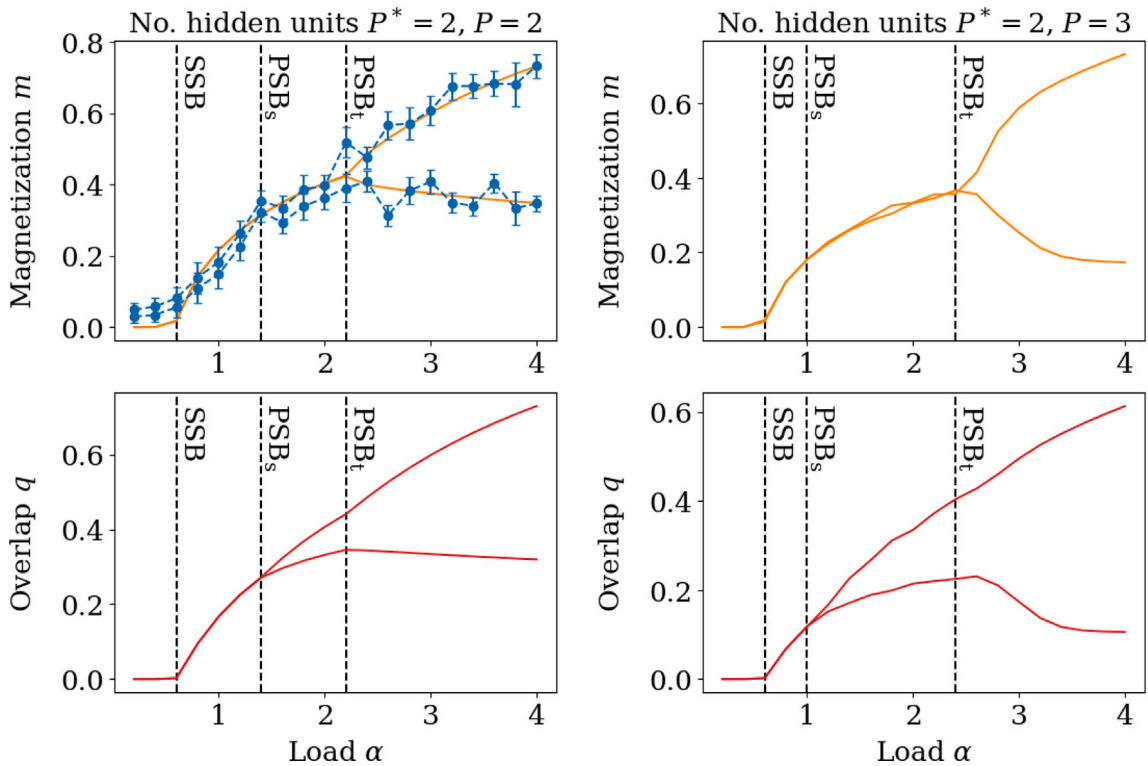

**Fig. 13.** Mattis magnetization $m$ and SG overlap $q$ solving Eqs. (14), in orange and red, as a function of the load $\alpha$ and number of student patterns $P$ for $\beta = \beta^* = 1$, $P^* = 2$ and $c = 0.3$. The top and bottom branches of the plots are respectively the diagonal and off-diagonal coefficients of $m$ and $q$. In the top-left panel, $m$ is compared against $N = 512$ dimensional Monte Carlo simulations, in blue. The blue dots and error bars represent the means and standard deviations, respectively, of the diagonal and off-diagonal coefficients of the magnetization $m$ during the simulations. Simulation results do not agree with the predictions of the saddle-point equations shown in the right panel when $\alpha > \alpha_{\text{crit}}$ (see Eq. (18)). Besides that, they are not very insightful, so we do not plot them for the sake of clarity.

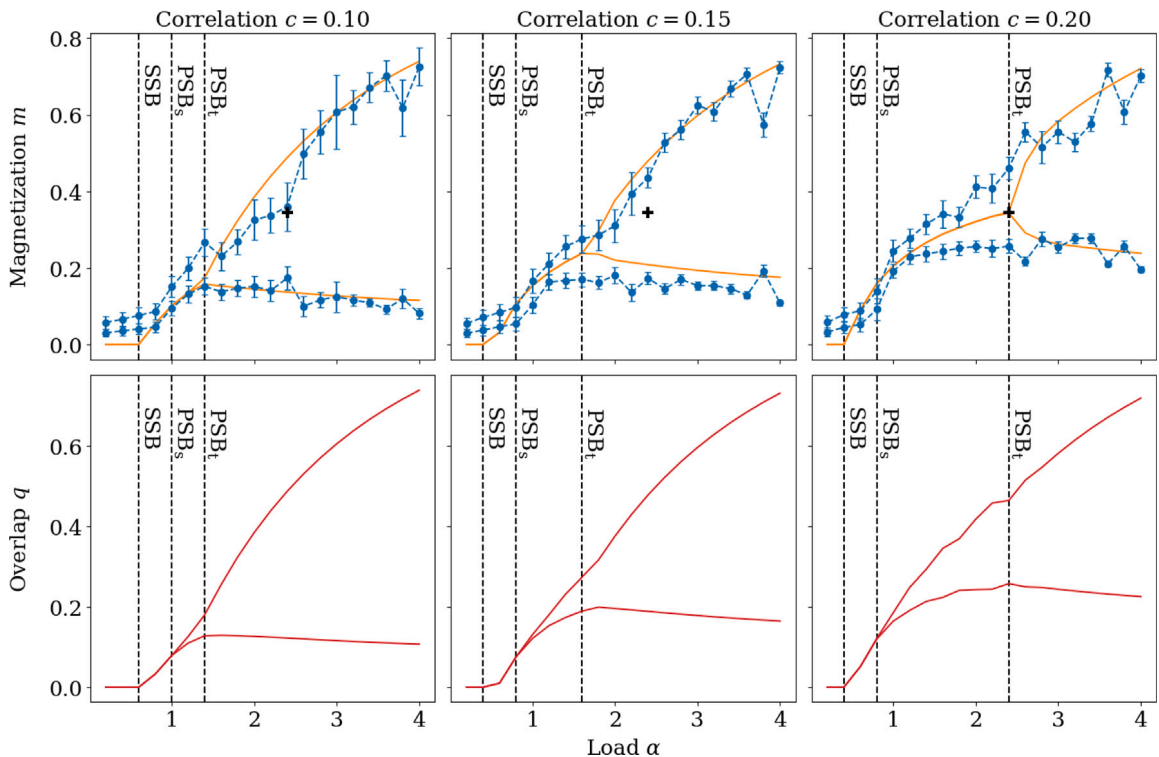

**Fig. 14.** Mattis magnetization $m$ and SG overlap $q$ solving Eqs. (14), in orange and red, as a function of the load $\alpha$ and teacher pattern correlations $c$ for $\beta = \beta^* = 1$ and $P = P^* = 3$. The top and bottom branches of the plots are respectively the diagonal and off-diagonal coefficients of $m$ and $q$. $m$ is compared against $N = 512$ dimensional Monte Carlo simulations, in blue. The blue dots and error bars represent the means and standard deviations, respectively, of the diagonal and off-diagonal coefficients of the magnetization $m$ during the simulations. The dashed lines indicate the approximate locations of the SSB, $\text{PSB}_s$ and $\text{PSB}_t$ phase transitions of Eqs. (14). The black crosses in the top plots mark the $\alpha$ and $m$ at which the $\text{PSB}_t$ transition occurs in the right plots. It serves as a visual guide to show that the $m$ predicted by Eqs. (14) can sometimes decrease significantly with $c$, but not the $m$ of the simulations.

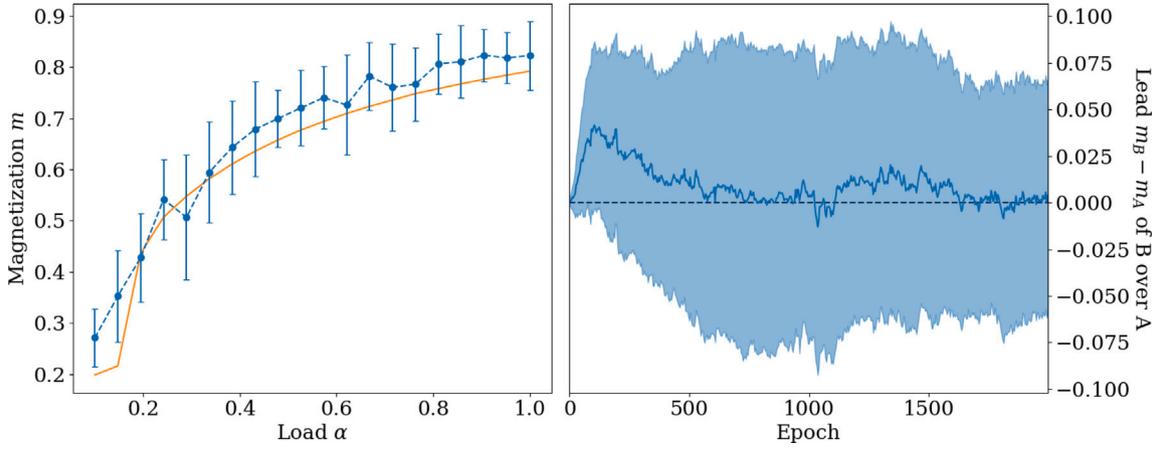

**Fig. 15.** Results of the lottery ticket experiment of Section 3.2.3 when the teacher patterns have a uniform correlation matrix with $c = 0.05$ instead of being i.i.d. In the left panel, $N = 512$ dimensional Monte Carlo simulations of student B, in blue, are compared against the solution of Eqs. (22), in orange. The blue dots and error bars represent the means and standard deviations, respectively, of the diagonal of the magnetization $m$ during the simulations. The right panel shows the difference $m_B − m_A$ of the magnetizations of A and B as a function of the simulation epochs. The solid blue line and the shaded region represent the median of $m_A − m_B$ over $\alpha \in [0, 1]$ and the corresponding mean absolute deviation around the median, respectively. $m_A − m_B$ goes to zero when the number of elapsed epochs is large, so student A converges to the solution of Eqs. (23) like student B. The inverse temperature is set to $\beta^* = \beta = 4$.

In Hou et al. (2019), the critical load $\alpha_{\text{crit}}^{\text{PSB}_t}$ increases with the correlation strength $c$. As such, increasing $c$ can trigger a phase transition from PSB$_t$ to PSB$_s$ and thus decrease the magnetization. In other words, large correlations in the teacher patterns can undermine the student's ability to learn them accurately. Based on these findings, Hou et al. Hou et al. (2019) formulated the hypothesis that the best $c$ for learning a relatively low load of data is non-zero but still relatively small. In the case of $P, P^* \geq 3$ (see Fig. 14), we also find that $\alpha_{\text{crit}}^{\text{PSB}_t}$ grows with $c$. However, unlike $P = P^* = 2$, Monte Carlo simulations of $P, P^* \geq 3$ do not completely agree with the RS approximation at large $c$. In particular, increasing $c$ does not seem to decrease the magnetization significantly in the simulations, even when it does so according to the RS approximation (see Fig. 14, black crosses).

Repeating the lottery ticket experiment of Section 3.2.3 for $c = 0.05$, we find that the lead of the winning ticket B over its randomly-initialized counterpart A has a smaller maximum (see Fig. 15). In other words, small correlations in the data appear to make the student less sensitive to initial conditions, which also reduces the benefits of magnitude pruning (see Section 3.2.3). As in Sections 3.2.2 and 3.2.3, the error in $m$ (see Fig. 15, left panel) is useful for measuring the convergence of the learning algorithm, but poorly representative of the equilibrium distribution.

Outside the teacher–student setting, RBMs typically undergo a sequence of second-order phase transitions at the beginning of training (Bachtis, Biroli, Decelle, & Seoane, 2024; Béreux, Decelle, Furtlehner, Rosset, & Seoane, 2024; Decelle et al., 2017, 2018). After the first transition, they learn a rank-one matrix, which is reminiscent of our SSB phase. The second transition then breaks the rank-one symmetry like our PSB$_s$ transition. The subsequent phase transitions are harder to interpret from the point of view of permutation symmetry breaking, and it is not clear whether a transition analogous to PSB$_t$ can occur without an explicit teacher. PSB$_s$ phase transitions were also observed in Gaussian mixture models (Akaho & Kappen, 2000; Decelle & Furtlehner, 2021; Kloppenburg & Tavan, 1997; Rose, Gurewitz, & Fox, 1990) and modern Hopfield networks (Boukacem et al., 2024), which suggests that they could be common in machine learning models.

### 3.4. Random correlations

In this section, we take the covariance matrix $Q$ to be random. By definition, $Q$ must be positive semi-definite. Moreover, when the teacher weights are binary, $Q$ must have ones on the diagonal. We sample $Q$ from the projected Wishart distribution $\mathcal{W}(c, D)$ defined in Appendix A.2 because it satisfies these two requirements. By the law of large numbers, $Q \sim \mathcal{W}(c, D)$ approaches $C = \delta_{\mu\nu} + (1 - \delta_{\mu\nu}) c$ in probability as $D \to \infty$ (see Appendix A.2). Therefore, $\lambda_{\max}^S$ and $\alpha_{\text{crit}}$ are the same as in Section 3.3 when $D$ is large. At finite $D = P$, the dependence of $\alpha_{\text{crit}}$ on $T$ is still qualitatively similar to that of uniform correlations (see Figs. 16 and J.21). However, the arithmetic mean $\overline{\alpha_{\text{crit}}}$ and the harmonic mean $\overline{\left[1/\lambda_{\max}^S\right]}^{-1}$ over many independent samples $Q \sim \mathcal{W}(c, P = D)$ are very different from the $\alpha_{\text{crit}}$ and $\lambda_{\max}^S$ of uniform correlations as a function of $c$ and $P$ (see Figs. 8 and 9). For instance, $\overline{\alpha_{\text{crit}}}$ decreases much more slowly with $c$ than the critical load of uniform correlations. At small $c$ and high $T$, $\overline{\alpha_{\text{crit}}}$ tends to be smaller than the critical load of uniform correlations. Conversely, at large $c$ and low $T$, the critical load of uniform correlations is usually smaller. At any given $c$, the entries of a random correlation matrix can sometimes be larger than $c$, which can make learning possible even when $T$ is too high for correlations of size $c$ or smaller to be picked up on by the student. However, this advantage disappears at larger $c$ and lower $T$ where correlations of size $c$ and smaller are no longer muddled in noise. In summary, the behavior of the critical load as a function of $P$ and $c$ is in general very different for random correlations and for uniform correlations.

### 4. Conclusion

In this paper, we theoretically study the learning performance of restricted Boltzmann machines (RBM) (Ackley et al., 1985; Freund & Haussler, 1991; Hinton, 2002; Smolensky, 1986) with a finite number of hidden units in the teacher–student setting (Barra et al., 2017, 2018; Decelle et al., 2021; Hou et al., 2019; Huang, 2017, 2018; Huang & Toyoizumi, 2016) using the replica method under the replica-symmetric (RS) approximation (Charbonneau, 2022). Given $M$ data points and $N$ visible units, we compute the critical data load $\alpha_{\text{crit}} = \frac{M}{N}$ above which learning becomes possible. Our findings validate the conjecture that the student's performance is independent of the number of hidden units when the patterns of the teacher are uncorrelated (Barra et al., 2017), generalizing the results of Hou et al. (2019) to any number of hidden units $P$ and teacher patterns $P^*$ much smaller than $M$ and $N$. In particular, we confirm that an RBM with $P$ uncorrelated hidden units factorizes into $P$ RBMs with one hidden unit each (Hou et al., 2019). Additionally, we show that the teacher–student setting without correlations has both a permutation symmetry breaking (PSB) solution in which the student learns the teacher patterns one-to-one and a metastable partial PSB solutions in which multiple student patterns can converge to the same teacher pattern. We argue that the teacher–student setting complies with the lottery ticket hypothesis (Frankle & Carbin, 2019; Frankle et al., 2020; Malach et al., 2020; Ramanujan et al., 2020; Zhou et al., 2019), and demonstrate that the student can be trained efficiently using a variant of the magnitude pruning algorithm (Frankle & Carbin, 2019; Frankle et al., 2020). Given teacher patterns with uniform correlations $c \in [0, 1)$, we find a closed-form expression for $\alpha_{\text{crit}}$, which generalizes the one found in Hou et al. (2019) for $P = P^* = 2$, and show that it decreases with both $c$ and $P^*$. Still in the case of uniform correlations, we find that the teacher–student setting undergoes a sequence of spontaneous symmetry breaking (SSB) and permutation symmetry breaking transitions that generalizes the one found in Hou et al. (2019) to $P^*, P \geq 3$. Both with and without correlations, we find that decreasing the inference temperature $T$ too much can prevent the student from learning the teacher's solution even when the dataset is relatively large. Throughout the paper, we compare key results against Monte Carlo simulations and observe that the RS ansatz is often a solid approximation even in the region outside the Nishimori line (Contucci et al., 2009; Iba, 1999; Nishimori, 1980, 2001) where the teacher and the student have the same temperature, but a different number of hidden units and a different prior on their patterns.

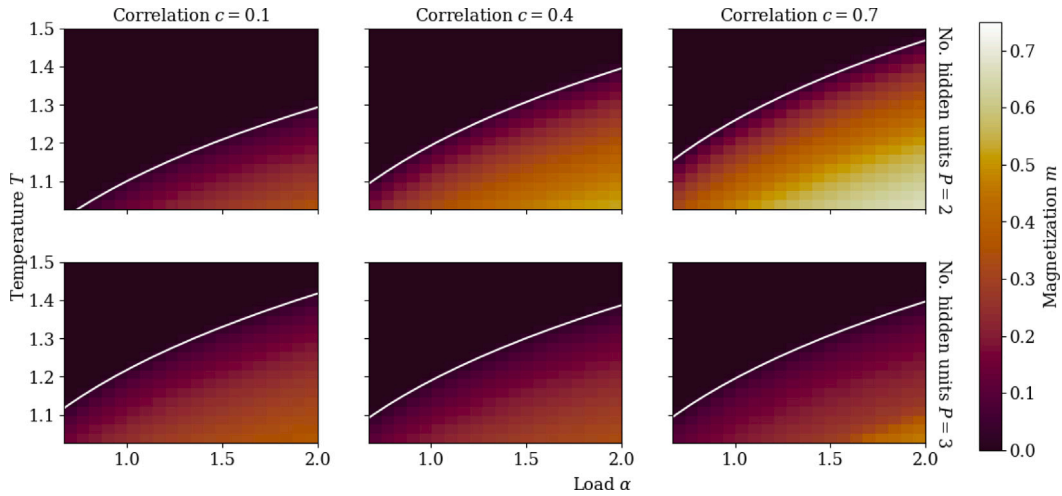

**Fig. 16.** Mattis magnetization $m$ for $\beta = \beta^*$ and $P = P^*$ as a function of the number of hidden units $P$, the correlation $c$, the temperature $T$ and the data load $\alpha$. $m$ is obtained by solving Eqs. (14) for binary student patterns with a uniform prior and binary teacher patterns with covariance $\mathcal{Q}_{\mu\nu} \sim \mathcal{W}(c, P)$, where $c \in [0, 1)$ (see Appendix A.2). The top and bottom rows feature $P = 2$ and $P = 3$, respectively. The white lines mark the phase transition described by Eq. (18) with $\lambda^S_{\max}$ given by Eq. (24).

One could study the learning dynamics of the magnitude pruning experiment described in Section 3.2.3 or characterize the distribution of winning ticket initial conditions to gain additional insight into the lottery ticket hypothesis. More generally, it would be interesting to extend our study to the case where $P$ and $P^*$ are of order $N$. In fact, RBMs trained on real data typically have a large number of hidden units (Fischer & Igel, 2014; Hinton, 2002; Kivinen & Williams, 2012; Le Roux et al., 2011; Tubiana & Monasson, 2017). One could also introduce correlations between the columns $\xi^*_i = \{\xi^{*\mu}_i\}^P_{\mu=1}$ of the teacher's weight matrix $\xi^* = \{\xi^{*\mu}_i\}^{1 \leq \mu \leq P}_{1 \leq i \leq N}$ in addition to the correlations between patterns $\xi^{*\mu} = \{\xi^{*\mu}_i\}^N_{i=1}$ that we study in this work. Such a modification could be achieved by using a weight matrix with a low-rank SVD structure like in Decelle et al. (2017, 2018) and would allow the teacher to generate data with even more structure than in this paper. One could use this framework to investigate whether the distinction between $\text{PSB}_s$ and $\text{PSB}_t$ persists in real data and has a measurable effect on the critical slowing down patterns observed in RBM training (Bachtis et al., 2024; Béreux et al., 2024; Decelle et al., 2017, 2018). Another promising research avenue would be to study other generative models with a finite number of hidden units in the teacher–student setting, such as modern (a.k.a. dense) Hopfield networks (Chen et al., 1986; Hopfield, 1982; Krotov & Hopfield, 2016; Psaltis & Park, 1986; Thériault & Tantari, 2024). In sum, the teacher–student setting still has a lot of untapped potential for studying generative models with many hidden units.

## CRediT authorship contribution statement

**Robin Thériault:** Writing – original draft, Validation, Software, Investigation, Formal analysis. **Francesco Tosello:** Software, Investigation, Formal analysis, Conceptualization. **Daniele Tantari:** Writing – review & editing, Writing – original draft, Validation, Supervision, Project administration, Investigation, Funding acquisition, Conceptualization.

## Declaration of competing interest

The authors declare that they have no known competing financial interests or personal relationships that could have appeared to influence the work reported in this paper.

## Acknowledgments

This work was partially supported by project SERICS (PE00000014) under the MUR National Recovery and Resilience Plan funded by the European Union - NextGenerationEU. The work was also supported by the project PRIN22TANTARI "Statistical Mechanics of Learning Machines: from algorithmic and information-theoretical limits to new biologically inspired paradigms" 20229T9EAT – CUP J53D23003640001. DT also acknowledges GNFM-Indam.

RT acknowledges the computational resources of the Center for High Performance Computing (CHPC) at the Scuola Normale Superiore of Pisa.

## Appendix A. Definitions

### A.1. Binary random variables with a fixed covariance matrix

We generate binary random variables $\xi_i^* \in \{-1, +1\}^{P^*}$ with mean 0 and a fixed covariance matrix $Q$ in two steps:

- sample $x_i^* \sim \mathcal{N}\left(0, \sin\left(\frac{\pi}{2}Q\right)\right)$ from a multivariate Gaussian distribution with mean 0 and covariance matrix $\sin\left(\frac{\pi}{2}Q\right)$,
- set $\xi_i^* = \text{sign}(x_i^*)$.

This sampling method is commonly known as the arcsine law and originates from (Van Vleck & Middleton, 1966). It implicitly defines a probability distribution $P(\xi^*)$ for correlated binary teacher patterns $\xi^*$.

### A.2. Projected wishart distribution

We generate random positive definite matrices $B$ by sampling them from the Wishart distribution (Wishart, 1928). That is, we sample the columns of a $P \times D$ matrix $\mathcal{A}$ from the Gaussian distribution $\mathcal{N}(0, C)$, then set $B = \mathcal{A}\mathcal{A}^T$. Let $D$ be the diagonal matrix with the same values as $B$ on the diagonal. We obtain a positive definite matrix $Q$ with ones on the diagonal by normalizing $B$ according to $Q = D^{-1/2} B D^{-1/2}$. In other words, $B$ is a generic covariance matrix, and the entries $Q_{\mu\nu}$ of $Q$ are the Pearson correlation coefficients corresponding to $B_{\mu\nu}$. For simplicity's sake, we take the covariance matrix $C$ of the Gaussian distribution $\mathcal{N}(0, C)$ to be $C_{\mu\nu} = \delta_{\mu\nu} + (1 - \delta_{\mu\nu})c$. In this paper, we say that $Q$ is sampled from the projected Wishart distribution $\mathcal{W}(c, D)$.

### A.3. Effective Hamiltonian $\mathcal{L}_{\lambda_1 \lambda_2}$

The effective Hamiltonian $\mathcal{L}_{\lambda_1 \lambda_2}$ plays a crucial role in deriving the free entropy and the saddle-point equations of the problem studied in this work. In this Section, we define it to be

$$\mathcal{L}_{\lambda_1, \lambda_2}(\xi, \xi^*, z; m, s, q) = \frac{1}{2}[\lambda_2]^2 \sum_{\mu,\nu=1}^{P} \left(s^{\mu\nu} - \frac{q^{\mu\nu} + q^{\nu\mu}}{2}\right) \xi^\mu \xi^\nu$$
$$+ \lambda_1 \lambda_2 \sum_{\gamma=1}^{P^*} \sum_{\mu=1}^{P} m^{\gamma\mu} \xi^{*\gamma} \xi^\mu$$
$$+ \lambda_2 \sum_{\mu,\nu=1}^{P} \sqrt{q^{\mu\nu} + q^{\nu\mu} - \delta_{\mu\nu} \sum_{\eta=1}^{P} \frac{q^{\mu\eta} + q^{\eta\mu}}{2}} z_{\mu\nu} \frac{\xi^\mu + \xi^\nu}{2}.$$

where we set the diagonal of $s$ to $s^{\mu\mu} = 0$ to regroup $s^{\mu\nu}$ and $\frac{q^{\mu\nu}+q^{\nu\mu}}{2}$ in the same sum. In $\mathcal{L}^O$ (see Eq. (9)), the diagonal of $s$ is arbitrary because it does not affect the saddle-point equations (see Eqs. (14)). In $\mathcal{L}^C$ (see Eq. (10)), the diagonal of $\hat{s}$ always vanishes (also see Eqs. (14)). When $q$ and $\hat{q}$ are symmetric, we obtain the simpler form

$$\mathcal{L}_{\lambda_1, \lambda_2}(\xi, \xi^*, z; m, s, q) = \frac{1}{2}[\lambda_2]^2 \sum_{\mu,\nu} (s^{\mu\nu} - q^{\mu\nu})\xi^\mu \xi^\nu + \lambda_1 \lambda_2 \sum_{\gamma,\mu} m^{\gamma\mu} \xi^{*\gamma} \xi^\mu$$
$$+ \lambda_2 \sum_{\mu,\nu} \sqrt{2q^{\mu\nu} - \delta_{\mu\nu} \sum_{\eta} q^{\mu\eta}} z_{\mu\nu} \frac{\xi^\mu + \xi^\nu}{2},$$

which is the definition used in the main text (see Eq. (11)). We symmetrize the initial conditions of $q^{\mu\nu}$ before solving the saddle-point equations numerically because any $q^{\mu\nu}$ solving them must be symmetric (see Eq. (14) and Appendix I). Defining

$$A_{\mu\nu}(q) = \sqrt{2q^{\mu\nu} - \delta_{\mu\nu} \sum_{\eta} q^{\mu\eta}},$$

these two expressions can be written more compactly as

$$\mathcal{L}_{\lambda_1, \lambda_2}(\xi, \xi^*, z; m, s, q) = \frac{1}{2}[\lambda_2]^2 \sum_{\mu,\nu} (s^{\mu\nu} - q^{\mu\nu})\xi^\mu \xi^\nu + \lambda_1 \lambda_2 \sum_{\gamma,\mu} m^{\gamma\mu} \xi^{*\gamma} \xi^\mu \quad (A.1)$$

$$+ \lambda_2 \sum_{\mu,\nu} A_{\mu\nu}\left(\frac{q + q^T}{2}\right) z_{\mu\nu} \frac{\xi^\mu + \xi^\nu}{2} \quad \text{for generic } q \text{ and}$$

$$\mathcal{L}_{\lambda_1, \lambda_2}(\xi, \xi^*, z; m, s, q) = \frac{1}{2}[\lambda_2]^2 \sum_{\mu,\nu} (s^{\mu\nu} - q^{\mu\nu})\xi^\mu \xi^\nu + \lambda_1 \lambda_2 \sum_{\gamma,\mu} m^{\gamma\mu} \xi^{*\gamma} \xi^\mu \quad (A.2)$$

$$+ \lambda_2 \sum_{\mu,\nu} A_{\mu\nu}(q) z_{\mu\nu} \frac{\xi^\mu + \xi^\nu}{2} \quad \text{for symmetric } q.$$

This way of writing $\mathcal{L}_{\lambda_1, \lambda_2}$ is useful to make the derivation of the saddle-point equations more concise (see Appendix D). The third term of $\mathcal{L}_{\lambda_1, \lambda_2}$ can also be written as

$$\lambda_2 \sum_{\mu,\nu} A_{\mu\nu}(q) z_{\mu\nu} \frac{\xi^\mu + \xi^\nu}{2} = \lambda_2 \sum_{\mu,\nu} A_{\mu\nu}(q) \frac{z_{\mu\nu} + z_{\nu\mu}}{2} \xi^\mu$$

by interchanging summation indices. The two ways of writing the third term have different numerical implementations, but we did not see a significant difference in computational complexity and accuracy between them.

## Appendix B. Replicated partition function

In the following appendices, we omit the subscripts $\beta^*$ and $\beta$ to make notation lighter. The sums over $\gamma$ or $\rho$ are from 1 to $P^*$, and the sums over $\mu$ or $\nu$, from 1 to $P$. Given a set of $M$ examples $\boldsymbol{\sigma} = \{\sigma^a\}_{a=1}^{M}$, the probability distribution of a single replica $b$ takes the form

$$P(\xi^b | \boldsymbol{\sigma}) = \mathcal{Z}(\boldsymbol{\sigma})^{-1} P(\xi^b) \prod_a P(\sigma^a | \xi^b),$$

by using Bayes' theorem (see Section 2). The partition function is $\mathcal{Z}(\boldsymbol{\sigma}) = \mathbb{E}_{\xi^b}\left[\prod_a P(\sigma^a | \xi^b)\right]$, which leads to

$$\mathbb{E}_{\xi^*, \boldsymbol{\sigma}}\left[\mathcal{Z}^L\right] = \sum_{\boldsymbol{\sigma}} \mathbb{E}_{\xi^*}\left[\prod_a P(\sigma^a | \xi^*)\right] \mathcal{Z}(\boldsymbol{\sigma})^L$$
$$= \sum_{\boldsymbol{\sigma}} \mathbb{E}_{\xi^*}\left[\prod_a P(\sigma^a | \xi^*)\right] \mathbb{E}_{\xi}\left[\prod_a P(\sigma^a | \xi)\right]$$
$$= \mathbb{E}_{\xi^*, \xi}\left[\prod_a \sum_{\sigma^a} P(\sigma^a | \xi^*) P(\sigma^a | \xi)\right]$$
$$= \mathbb{E}_{\xi^*, \xi}\left[\left(\sum_{\sigma^a} P(\sigma^a | \xi^*) P(\sigma^a | \xi)\right)^M\right]$$

where $\xi = \{\xi^b\}_{b=1}^L$ is a set of $L$ replicas and $P(\sigma^a | \xi) = \prod_b P(\sigma^a | \xi^b)$. As per Eq. (2), $P(\sigma^a | \xi^b)$ with binary units takes the form

$$P(\sigma^a | \xi^b) = Z(\xi^b)^{-1} \psi(\sigma^a; \xi^b),$$

where $\psi(\sigma^a; \xi^b) = P(\sigma^a) \mathbb{E}_{\tau_b}\left[\exp\left(\frac{\beta}{\sqrt{N}} \sum_\mu \tau_{b\mu} \sum_i \xi_i^{b\mu} \sigma_i^a\right)\right]$ and $Z(\xi^b) = \sum_{\sigma^a} \psi(\sigma^a; \xi^b)$. In particular, we have

$$\sum_{\sigma^a} P(\sigma^a | \xi^*) P(\sigma^a | \xi) = \sum_{\sigma^a} Z(\xi^*)^{-1} \left[\prod_b Z(\xi^b)^{-1}\right] \psi(\sigma^a; \xi^*) \left[\prod_b \psi(\sigma^a; \xi^b)\right].$$

We will now take $P(\sigma^a)$ uniform and eliminate $\sigma^a$ from $Z(\xi^*) \prod_b Z(\xi^b)$ in order to factor it out of the sum over $\sigma^a$. We first rewrite $Z(\xi^b)$ as

$$Z(\xi^b) = 2^{-N} \sum_{\sigma^a} \mathbb{E}_{\tau_b}\left[\exp\left(\frac{\beta}{\sqrt{N}} \sum_\mu \tau_{b\mu} \sum_i \xi_i^{b\mu} \sigma_i^a\right)\right]$$
$$= 2^{-N} \mathbb{E}_{\tau_b}\left[\prod_i \sum_{\sigma_i^a} \exp\left(\frac{\beta}{\sqrt{N}} \sum_\mu \tau_{b\mu} \xi_i^{b\mu} \sigma_i^a\right)\right]$$
$$= \mathbb{E}_{\tau_b}\left[\prod_i \cosh\left(\frac{\beta}{\sqrt{N}} \sum_\mu \tau_{b\mu} \xi_i^{b\mu}\right)\right]$$
$$= \mathbb{E}_{\tau_b}\left[\exp\left(\sum_i \log \cosh\left[\frac{\beta}{\sqrt{N}} \sum_\mu \tau_{b\mu} \xi_i^{b\mu}\right]\right)\right],$$

then we expand the log cosh function in small $\frac{P}{N}$ to obtain

$$Z\left(\xi^{b}\right) \approx \mathbb{E}_{\tau_{b}}\left[\exp \left(\sum_{i} \frac{1}{2}\left[\frac{\beta}{\sqrt{N}} \sum_{\mu} \tau_{b\mu} \xi_{i}^{b\mu}\right]^{2}\right)\right]$$

$$= \mathbb{E}_{\tau_{b}}\left[\exp \left(\frac{1}{2} \beta^{2} \sum_{\mu,\nu} \tau_{b\mu} \tau_{b\nu} \frac{1}{N} \sum_{i} \xi_{i}^{b\mu} \xi_{i}^{b\nu}\right)\right].$$

The equivalent expression for $Z(\xi^*)$ is identical except for the asterisk $*$ replacing the replica index $b$ and the sum running from 1 to $P^*$ rather than 1 to $P$. Once $Z(\xi^*) \prod_b Z(\xi^b)$ is factored out of the sum, $\sum_{\sigma^a} \psi(\sigma^a; \xi^*) \prod_b \psi(\sigma^a; \xi^b)$ simplifies in a similar way to

$$\sum_{\sigma^{a}} \psi\left(\sigma^{a} ; \xi^{*}\right) \prod_{b} \psi\left(\sigma^{a} ; \xi^{b}\right) \approx \mathbb{E}_{\tau_{*},\tau}\left[\exp \left(\sum_{i} \frac{1}{2}\left[\frac{\beta^{*}}{\sqrt{N}} \sum_{\gamma} \tau_{*\gamma} \xi_{i}^{*\gamma} + \frac{\beta}{\sqrt{N}} \sum_{\mu;b} \tau_{b\mu} \xi_{i}^{b\mu}\right]^{2}\right)\right]$$

$$= \mathbb{E}_{\tau_{*},\tau}\left[\exp \left(\frac{1}{2}\left[\beta^{*}\right]^{2} \sum_{\gamma,\rho} \tau_{*\gamma} \tau_{*\rho} \frac{1}{N} \sum_{i} \xi_{i}^{*\gamma} \xi_{i}^{*\rho}\right. \right.$$
$$\left. + \beta^{*} \beta \sum_{\gamma,\mu;b} \tau_{*\gamma} \tau_{b\mu} \frac{1}{N} \sum_{i} \xi_{i}^{*\gamma} \xi_{i}^{b\mu}$$
$$\left. + \frac{1}{2} \beta^{2} \sum_{\mu,\nu;b,c} \tau_{b\mu} \tau_{c\nu} \frac{1}{N} \sum_{i} \xi_{i}^{b\mu} \xi_{i}^{c\nu}\right)\right],$$

up to an irrelevant multiplicative factor of $2^{-LN}$. Combining all these expressions, we get

$$\mathbb{E}\left[\mathcal{Z}^{L}\right] = \mathbb{E}_{\xi^{*}\xi}\left[\exp \left\{M\left(\log \left[\mathbb{E}_{\tau_{*},\tau} \exp \left(\left[\beta^{*}\right]^{2} \sum_{\gamma<\rho} \tau_{*\gamma} \tau_{*\rho} \frac{1}{N} \sum_{i} \xi_{i}^{*\gamma} \xi_{i}^{*\rho}\right.\right.\right.\right.\right.$$
$$+ \beta^{*} \beta \sum_{\gamma,\mu;b} \tau_{*\gamma} \tau_{b\mu} \frac{1}{N} \sum_{i} \xi_{i}^{*\gamma} \xi_{i}^{b\mu} + \beta^{2} \sum_{\mu<\nu;b} \tau_{b\mu} \tau_{b\nu} \frac{1}{N} \sum_{i} \xi_{i}^{b\mu} \xi_{i}^{b\nu}$$
$$+ \beta^{2} \sum_{\mu,\nu;b<c} \tau_{b\mu} \tau_{c\nu} \frac{1}{N} \sum_{i} \xi_{i}^{b\mu} \xi_{i}^{c\nu}\bigg)\bigg]$$
$$- \log \left[\mathbb{E}_{\tau_{*}} \exp \left(\left[\beta^{*}\right]^{2} \sum_{\gamma<\rho} \tau_{*\gamma} \tau_{*\rho} \frac{1}{N} \sum_{i} \xi_{i}^{*\gamma} \xi_{i}^{*\rho}\right)\right]$$
$$- \log \left[\mathbb{E}_{\tau} \exp \left(\beta^{2} \sum_{\mu<\nu;b} s^{b\mu\nu} \tau_{b\mu} \tau_{b\nu}\right)\right]\bigg)\bigg\}\bigg].$$

## Appendix C. RS free entropy

In this Section, we introduce order parameters and conjugate order parameters for the various overlaps between patterns present in $\mathbb{E}[\mathcal{Z}^L]$. We define

$m^{b\gamma\mu}$ and $\hat{m}^{b\gamma\mu}$ for $\frac{1}{N} \sum_{i} \xi_{i}^{*\gamma} \xi_{i}^{b\mu}$,

$s^{b\mu\nu}$ and $\hat{s}^{b\mu\nu}$ for $\frac{1}{N} \sum_{i} \xi_{i}^{b\mu} \xi_{i}^{b\nu}$ where $\mu \neq \nu$,

$q^{bc\mu\nu}$ and $\hat{q}^{bc\mu\nu}$ for $\frac{1}{N} \sum_{i} \xi_{i}^{b\mu} \xi_{i}^{c\nu}$ where $b \neq c$,

where $m^{b\gamma\mu}$, $s^{b\mu\nu}$ and $q^{bc\mu\nu}$ are the ordinary order parameters, $\hat{m}^{b\gamma\mu}$, $\hat{s}^{b\mu\nu}$ and $\hat{q}^{bc\mu\nu}$ are the conjugate order parameters and $s^{b\mu\nu}$ and $\hat{s}^{b\mu\nu}$ are symmetric in $\mu$ and $\nu$ by construction. We use a Fourier transform to rewrite $\mathbb{E}[\mathcal{Z}^L]$ as

$$\mathbb{E}\left[\mathcal{Z}^{L}\right] = \mathbb{E}_{\xi^{*}\xi}\left[\int \prod_{i\mathbb{R}} \prod_{\gamma,\mu;b} d\hat{m}^{b\gamma\mu} \prod_{\mu<\nu;b} d\hat{s}^{b\mu\nu} \prod_{\mu,\nu;b<c} d\hat{q}^{bc\mu\nu}\right.$$
$$\int_{\mathbb{R}} \prod_{\gamma,\mu;b} dm^{b\gamma\mu} \prod_{\mu<\nu;b} ds^{b\mu\nu} \prod_{\mu,\nu;b<c} dq^{bc\mu\nu} \exp \left\{\sum_{\gamma,\mu;b} \hat{m}^{b\gamma\mu} \left(\sum_{i} \xi_{i}^{*\gamma} \xi_{i}^{b\mu} - Nm^{b\gamma\mu}\right)\right\}$$
$$\exp \left\{\sum_{\mu<\nu;b} \hat{s}^{b\mu\nu} \left(\sum_{i} \xi_{i}^{b\mu} \xi_{i}^{b\nu} - Ns^{b\mu\nu}\right) + \sum_{\mu,\nu;b<c} \hat{q}^{bc\mu\nu} \left(\sum_{i} \xi_{i}^{b\mu} \xi_{i}^{c\nu} - Nq^{bc\mu\nu}\right)\right\}$$
$$\exp \left\{M\left(\log \left[\mathbb{E}_{\tau_{*},\tau} \exp \left(\left[\beta^{*}\right]^{2} \sum_{\gamma<\rho} \tau_{*\gamma} \tau_{*\rho} \frac{1}{N} \sum_{i} \xi_{i}^{*\gamma} \xi_{i}^{*\rho}\right.\right.\right.\right.$$
$$+ \beta^{*} \beta \sum_{\gamma,\mu;b} m^{b\gamma\mu} \tau_{*\gamma} \tau_{b\mu} + \beta^{2} \sum_{\mu<\nu;b} s^{b\mu\nu} \tau_{b\mu} \tau_{b\nu} + \beta^{2} \sum_{\mu,\nu;b<c} q^{bc\mu\nu} \tau_{b\mu} \tau_{c\nu}\bigg)\bigg]$$

$$- \log \left[\mathbb{E}_{\tau_{*}} \exp \left(\left[\beta^{*}\right]^{2} \sum_{\gamma<\rho} \tau_{*\gamma} \tau_{*\rho} \frac{1}{N} \sum_{i} \xi_{i}^{*\gamma} \xi_{i}^{*\rho}\right)\right]$$
$$- \log \left[\mathbb{E}_{\tau} \exp \left(\beta^{2} \sum_{\mu<\nu;b} s^{b\mu\nu} \tau_{b\mu} \tau_{b\nu}\right)\right]\bigg)\bigg\}\bigg].$$

By hypothesis, the columns $\xi_{i}^{*} = \{\xi_{i}^{*\gamma}\}_{\gamma=1}^{P^{*}}$ of $\xi^{*} \sim \mathrm{P}(\xi^{*})$ are i.i.d. random variables and their distribution $\mathrm{P}(\xi_{i}^{*})$ has a well-defined $P^{*} \times P^{*}$ covariance matrix $\mathcal{Q}$ (see Section 1). Therefore, by the law of large numbers, $\frac{1}{N} \sum_{i} \xi_{i}^{*\gamma} \xi_{i}^{*\rho}$ converges in probability to the covariance $\mathcal{Q}_{\gamma\rho}$ as $N \to \infty$. $\mathbb{E}[\mathcal{Z}^L]$ then simplifies to

$$\mathbb{E}\left[\mathcal{Z}^{L}\right] = \int \prod_{\gamma,\mu;b} d\hat{m}^{b\gamma\mu} dm^{b\gamma\mu} \prod_{\mu<\nu;b} d\hat{s}^{b\mu\nu} ds^{b\mu\nu} \prod_{\mu,\nu;b<c} d\hat{q}^{bc\mu\nu} dq^{bc\mu\nu}$$
$$\exp \left\{N \log \left[\mathbb{E}_{\xi_{i}^{*}\xi_{i}} \exp \left\{H_{S}\left(\xi_{i}, \xi_{i}^{*}; \hat{m}, \hat{s}, \hat{q}\right)\right\}\right] - NH_{Q}(m, s, q, \hat{m}, \hat{s}, \hat{q})\right\}$$
$$\exp \left\{\alpha N \log \left[\left\langle \mathbb{E}_{\tau} \exp \left\{H_{E}\left(\tau, \tau_{*}; m, s, q\right)\right\}\right\rangle_{\mathcal{M}_{*}}\right] - \alpha N \log [\mathcal{Z}(\mathcal{M})]\right\}$$

where the thermal average $\langle \cdot \rangle_{\mathcal{M}_{*}}$ and the partition function $\mathcal{Z}(\mathcal{M})$ are defined in Section 3.1 using Eqs. (7) and (8), respectively, $\alpha = \frac{M}{N}$ and

$$H_{Q}(m, s, q, \hat{m}, \hat{s}, \hat{q}) = \sum_{\gamma,\mu;b} \hat{m}^{b\gamma\mu} m^{b\gamma\mu} + \sum_{\mu<\nu;b} \hat{s}^{b\mu\nu} s^{b\mu\nu} + \sum_{\mu,\nu;b<c} \hat{q}^{bc\mu\nu} q^{bc\mu\nu},$$

$$H_{S}(\xi_{i}, \xi_{i}^{*}; \hat{m}, \hat{s}, \hat{q}) = \sum_{\gamma,\mu;b} \hat{m}^{b\gamma\mu} \xi_{i}^{*\gamma} \xi_{i}^{b\mu} + \sum_{\mu<\nu;b} \hat{s}^{b\mu\nu} \xi_{i}^{b\mu} \xi_{i}^{b\nu} + \sum_{\mu,\nu;b<c} \hat{q}^{bc\mu\nu} \xi_{i}^{b\mu} \xi_{i}^{c\nu},$$

$$H_{E}(\tau, \tau_{*}; m, s, q) = \beta^{*} \beta \sum_{\gamma,\mu;b} m^{b\gamma\mu} \tau_{*\gamma} \tau_{b\mu}$$
$$+ \beta^{2} \sum_{\mu<\nu;b} s^{b\mu\nu} \tau_{b\mu} \tau_{b\nu} + \beta^{2} \sum_{\mu,\nu;b<c} q^{bc\mu\nu} \tau_{b\mu} \tau_{c\nu}.$$

We use the replica symmetry ansatz to simplify $\mathbb{E}[\mathcal{Z}^L]$ further. To be more precise, we assume that

$m^{b\gamma\mu} = m^{\gamma\mu}$ and $\hat{m}^{b\gamma\mu} = \hat{m}^{\gamma\mu}$ for all $b; \gamma, \mu$,

$s^{b\mu\nu} = s^{\mu\nu}$ and $\hat{s}^{b\mu\nu} = \hat{s}^{\mu\nu}$ for all $b; \mu \neq \nu$,

$q^{bc\mu\nu} = q^{\mu\nu}$ and $\hat{q}^{bc\mu\nu} = \hat{q}^{\mu\nu}$ for all $b \neq c; \mu, \nu$.

Under this hypothesis, the term $\sum_{\mu,\nu;b<c} q^{\mu\nu} \tau_{b\mu} \tau_{c\nu}$ can be rewritten as

$$\sum_{\mu,\nu;b<c} q^{\mu\nu} \tau_{b\mu} \tau_{c\nu} = \frac{1}{2} \sum_{\mu,\nu;b,c} q^{\mu\nu} \tau_{b\mu} \tau_{c\nu} - \frac{1}{2} \sum_{\mu,\nu;b} q^{\mu\nu} \tau_{b\mu} \tau_{b\nu}$$

$$= \frac{1}{4} \sum_{\mu,\nu} q^{\mu\nu} \left[\sum_{b}\left(\tau_{b\mu} + \tau_{b\nu}\right)\right]^{2} - \frac{1}{4} \sum_{\mu,\nu} q^{\mu\nu} \left[\sum_{b} \tau_{b\mu}\right]^{2}$$

$$- \frac{1}{4} \sum_{\mu,\nu} q^{\mu\nu} \left[\sum_{b} \tau_{b\nu}\right]^{2} - \frac{1}{2} \sum_{\mu,\nu;b} q^{\mu\nu} \tau_{b\mu} \tau_{b\nu}$$

$$= \sum_{\mu,\nu} \frac{q^{\mu\nu} + q^{\nu\mu}}{2} \left[\sum_{b} \frac{\tau_{b\mu} + \tau_{b\nu}}{2}\right]^{2} - \frac{1}{2} \sum_{\mu,\nu} \frac{q^{\mu\nu} + q^{\nu\mu}}{2} \left[\sum_{b} \tau_{b\mu}\right]^{2}$$

$$- \frac{1}{2} \sum_{\mu,\nu;b} q^{\mu\nu} \tau_{b\mu} \tau_{b\nu}$$

$$= \frac{1}{2} \sum_{\mu,\nu} \left[q^{\mu\nu} + q^{\nu\mu} - \delta_{\mu\nu} \sum_{\eta} \frac{q^{\mu\eta} + q^{\eta\mu}}{2}\right] \left[\sum_{b} \frac{\tau_{b\mu} + \tau_{b\nu}}{2}\right]^{2}$$

$$- \frac{1}{2} \sum_{\mu,\nu;b} q^{\mu\nu} \tau_{b\mu} \tau_{b\nu}.$$

Subsequently, the Hubbard–Stratonovich transformation gives

$$H_{E}(\tau, \tau_{*}; m, s, q) = \beta^{*} \beta \sum_{\gamma,\mu;b} m^{\gamma\mu} \tau_{*\gamma} \tau_{b\mu} + \beta^{2} \sum_{\mu<\nu;b} s^{\mu\nu} \tau_{b\mu} \tau_{b\nu}$$
$$+ \beta^{2} \sum_{\mu,\nu;b<c} q^{\mu\nu} \tau_{b\mu} \tau_{c\nu}$$
$$= \log \left(\mathbb{E}_{z}\left[\prod_{b} \exp \left\{\mathcal{L}_{\beta^{*},\beta}\left(\tau_{b}, \tau_{*}, z; m, s, q\right)\right\}\right]\right)$$

where $\mathcal{L}_{\lambda_1,\lambda_2}(\xi, \xi^*, z; m, s, q)$ is defined in Appendix A.3 and $z = [z_{\mu\nu}]_{\mu,\nu=1}^{P}$ is a matrix of independent standard Gaussian random variables $z_{\mu\nu}$. Similarly, we get

$$H_{S}(\xi, \xi^{*}; \hat{m}, \hat{s}, \hat{q}) = \sum_{\gamma,\mu;b} \hat{m}^{\gamma\mu} \xi^{*\gamma} \xi^{b\mu} + \sum_{\mu<\nu;b} \hat{s}^{\mu\nu} \xi^{b\mu} \xi^{b\nu} + \sum_{\mu,\nu;b<c} \hat{q}^{\mu\nu} \xi^{b\mu} \xi^{c\nu}$$

$$= \log\left(\mathbb{E}_z\left[\prod_b \exp\left\{\mathcal{L}_{1,1}\left(\xi^b, \xi^*, z; \hat{m}, \hat{s}, \hat{q}\right)\right\}\right]\right).$$

We then factor $\mathbb{E}\left[\mathcal{Z}^L\right]$ over the replicas and take the limit of $L \to 0, N \to \infty$ to obtain

$$f = \text{Extr}_{m,\hat{m},q,\hat{q},s,\hat{s}} f(m, \hat{m}, q, \hat{q}, s, \hat{s}) \qquad (\text{C}.1)$$

$$= \text{Extr}_{m,\hat{m},q,\hat{q},s,\hat{s}} \Big\{ -\sum_{\gamma,\mu} m^{\gamma\mu} \hat{m}^{\gamma\mu} - \frac{1}{2}\sum_{\mu\ne\nu} s^{\mu\nu}\hat{s}^{\mu\nu} + \frac{1}{2}\sum_{\mu,\nu} q^{\mu\nu}\hat{q}^{\mu\nu}$$

$$+ \mathbb{E}_{\xi^*}\mathbb{E}_z\log\left[\mathcal{Z}\left(\mathcal{L}^C\right)\right] + \alpha\left\langle\mathbb{E}_z\log\left[\mathcal{Z}\left(\mathcal{L}^O\right)\right]\right\rangle_{\mathcal{M}_*} - \alpha\log\left[\mathcal{Z}\left(\mathcal{M}\right)\right]\Big\},$$

where we used the replica trick $\lim_{L\to 0}\left(\frac{1}{L}\log\mathbb{E}_z\left[\mathcal{Z}\left(\mathcal{L}_{\lambda_1,\lambda_2}\right)^L\right]\right) = \mathbb{E}_z\log\left[\mathcal{Z}\left(\mathcal{L}_{\lambda_1,\lambda_2}\right)\right]$ to simplify the expectations over $z$. The partition functions $\mathcal{Z}\left(\mathcal{L}^O\right)$ and $\mathcal{Z}\left(\mathcal{L}^C\right)$ are defined in Section 3.1 using Eqs. (9) and (10), respectively.

**Appendix D. Saddle-point equations**

Our goal is to find the values of the order parameters for which the derivatives of $f(m, \hat{m}, q, \hat{q}, s, \hat{s})$ vanish (see Eq. C). For that purpose, we need to evaluate

$$\partial_{m^{\rho\iota}}\log\left[\mathcal{Z}\left(\mathcal{L}_{\lambda_1,\lambda_2}\right)\right] = \left\langle\partial_{m^{\rho\iota}}\mathcal{L}_{\lambda_1,\lambda_2}\right\rangle_{\mathcal{L}_{\lambda_1,\lambda_2}}$$

$$\partial_{s^{\iota\kappa}}\log\left[\mathcal{Z}\left(\mathcal{L}_{\lambda_1,\lambda_2}\right)\right] = \left\langle\partial_{s^{\iota\kappa}}\mathcal{L}_{\lambda_1,\lambda_2}\right\rangle_{\mathcal{L}_{\lambda_1,\lambda_2}}$$

$$\partial_{q^{\iota\kappa}}\log\left[\mathcal{Z}\left(\mathcal{L}_{\lambda_1,\lambda_2}\right)\right] = \left\langle\partial_{q^{\iota\kappa}}\mathcal{L}_{\lambda_1,\lambda_2}\right\rangle_{\mathcal{L}_{\lambda_1,\lambda_2}},$$

so we must calculate the partial derivatives inside the expectation values. We use Eq. A.3 for $\mathcal{L}_{\lambda_1,\lambda_2}$ because we do not know that $q$ is symmetric before deriving the saddle-point equations (see Appendix A.3). We obtain

$$\partial_{m^{\rho\iota}}\mathcal{L}_{\lambda_1,\lambda_2} = \lambda_1\lambda_2 \xi^{*\rho}\xi^{\iota}$$

$$\partial_{s^{\iota\kappa}}\mathcal{L}_{\lambda_1,\lambda_2} = \frac{1}{2}\left[\lambda_2\right]^2 \xi^{\iota}\xi^{\kappa}$$

$$\partial_{q^{\iota\kappa}}\mathcal{L}_{\lambda_1,\lambda_2} = \frac{1}{4}\lambda_2\left(A_{\iota\kappa}^{-1}\left[z_{\iota\kappa}+z_{\kappa\iota}\right]\left[\xi^{\iota}+\xi^{\kappa}\right] - A_{\iota\iota}^{-1}z_{\iota\iota}\xi^{\iota} - A_{\kappa\kappa}^{-1}z_{\kappa\kappa}\xi^{\kappa}\right)$$
$$- \frac{1}{2}\left[\lambda_2\right]^2 \xi^{\iota}\xi^{\kappa},$$

where $A_{\mu\nu}$ is defined in Appendix A.3. Using the first two equalities, we immediately get

$$\hat{m}^{\gamma\mu} = \beta^*\beta\alpha \left\langle\mathbb{E}_z\left[\tau_{*\gamma}\left\langle\tau_\mu\right\rangle_{\mathcal{L}^O}\right]\right\rangle_{\mathcal{M}_*}$$

$$\hat{s}^{\mu\nu} = \beta^2\alpha\left(\left\langle\mathbb{E}_z\left[\left\langle\tau_\mu\tau_\nu\right\rangle_{\mathcal{L}^O}\right]\right\rangle_{\mathcal{M}_*} - \left\langle\tau_\mu\tau_\nu\right\rangle_{\mathcal{M}}\right)$$

$$m^{\gamma\mu} = \mathbb{E}_{\xi^*}\mathbb{E}_z\left[\xi^{*\gamma}\left\langle\xi^\mu\right\rangle_{\mathcal{L}^C}\right]$$

$$s^{\mu\nu} = \mathbb{E}_{\xi^*}\mathbb{E}_z\left[\left\langle\xi^\mu\xi^\nu\right\rangle_{\mathcal{L}^C}\right],$$

where the thermal averages $\langle\cdot\rangle_{\mathcal{M}_*}$, $\langle\cdot\rangle_{\mathcal{M}}$, $\langle\cdot\rangle_{\mathcal{L}^O}$ and $\langle\cdot\rangle_{\mathcal{L}^C}$ are defined in Section 3.1 using Eqs. (7), (8), (9) and (10), respectively. The case of the third derivative is a bit more involved. We find that its expectation with respect to the Gaussian variables $z$ can be expressed as

$$\mathbb{E}_z\left[\left\langle\partial_{q^{\iota\kappa}}\mathcal{L}_{\lambda_1,\lambda_2}\right\rangle_{\mathcal{L}_{\lambda_1,\lambda_2}}\right] = \mathbb{E}_z\Big[\frac{1}{4}\lambda_2\Big(A_{\iota\kappa}^{-1}\left[z_{\iota\kappa}+z_{\kappa\iota}\right]\langle\xi^\iota+\xi^\kappa\rangle_{\mathcal{L}_{\lambda_1,\lambda_2}} - A_{\iota\iota}^{-1}z_{\iota\iota}\langle\xi^\iota\rangle_{\mathcal{L}_{\lambda_1,\lambda_2}}$$
$$- A_{\kappa\kappa}^{-1}z_{\kappa\kappa}\langle\xi^\kappa\rangle_{\mathcal{L}_{\lambda_1,\lambda_2}}\Big) - \frac{1}{2}\left[\lambda_2\right]^2\langle\xi^\iota\xi^\kappa\rangle_{\mathcal{L}_{\lambda_1,\lambda_2}}\Big]$$

$$= \mathbb{E}_z\Big[\frac{1}{4}\lambda_2\Big(A_{\iota\kappa}^{-1}\left[\partial_{z_{\iota\kappa}}\langle\xi^\iota\rangle_{\mathcal{L}_{\lambda_1,\lambda_2}} + \partial_{z_{\kappa\iota}}\langle\xi^\iota\rangle_{\mathcal{L}_{\lambda_1,\lambda_2}}$$
$$+ \partial_{z_{\iota\kappa}}\langle\xi^\kappa\rangle_{\mathcal{L}_{\lambda_1,\lambda_2}} + \partial_{z_{\kappa\iota}}\langle\xi^\kappa\rangle_{\mathcal{L}_{\lambda_1,\lambda_2}}\Big] - A_{\iota\iota}^{-1}\partial_{z_{\iota\iota}}\langle\xi^\iota\rangle_{\mathcal{L}_{\lambda_1,\lambda_2}}$$
$$- A_{\kappa\kappa}^{-1}\partial_{z_{\kappa\kappa}}\langle\xi^\kappa\rangle_{\mathcal{L}_{\lambda_1,\lambda_2}}\Big) - \frac{1}{2}\left[\lambda_2\right]^2\langle\xi^\iota\xi^\kappa\rangle_{\mathcal{L}_{\lambda_1,\lambda_2}}\Big]$$

by using integration by parts. As known from linear response theory (Chandler, 1987), any Gibbs distribution with a generic Hamiltonian $\mathcal{H}$ verifies the identity

$$\partial_x\langle\theta\rangle_{\mathcal{H}} = \langle\theta\partial_x\mathcal{H}\rangle_{\mathcal{H}} - \langle\theta\rangle_{\mathcal{H}}\langle\partial_x\mathcal{H}\rangle_{\mathcal{H}} \qquad (\text{D}.1)$$

for any order parameter $\theta$ that does not depend on $x$. We use this identity to get

$$\partial_{z_{\iota\kappa}}\langle\xi^\iota\rangle_{\mathcal{L}_{\lambda_1,\lambda_2}} = \left\langle\xi^\iota\partial_{z_{\iota\kappa}}\mathcal{L}_{\lambda_1,\lambda_2}\right\rangle_{\mathcal{L}_{\lambda_1,\lambda_2}} - \left\langle\xi^\iota\right\rangle_{\mathcal{L}_{\lambda_1,\lambda_2}}\left\langle\partial_{z_{\iota\kappa}}\mathcal{L}_{\lambda_1,\lambda_2}\right\rangle_{\mathcal{L}_{\lambda_1,\lambda_2}}$$

$$= \lambda_2\left\langle A_{\iota\kappa}\xi^\iota\frac{\xi^\iota+\xi^\kappa}{2}\right\rangle_{\mathcal{L}_{\lambda_1,\lambda_2}} - \lambda_2\langle\xi^\iota\rangle_{\mathcal{L}_{\lambda_1,\lambda_2}}\left\langle A_{\iota\kappa}\frac{\xi^\iota+\xi^\kappa}{2}\right\rangle_{\mathcal{L}_{\lambda_1,\lambda_2}}$$

$$= \frac{1}{2}\lambda_2 A_{\iota\kappa}\left(\langle\xi^\iota\xi^\kappa\rangle_{\mathcal{L}_{\lambda_1,\lambda_2}} - \langle\xi^\iota\rangle_{\mathcal{L}_{\lambda_1,\lambda_2}}\langle\xi^\kappa\rangle_{\mathcal{L}_{\lambda_1,\lambda_2}}\right)$$

$$+ \frac{1}{2}\lambda_2 A_{\iota\kappa}\left(\left\langle\left[\xi^\iota\right]^2\right\rangle_{\mathcal{L}_{\lambda_1,\lambda_2}} - \left[\langle\xi^\iota\rangle_{\mathcal{L}_{\lambda_1,\lambda_2}}\right]^2\right),$$

Coming back to the previous expression, we obtain

$$\mathbb{E}_z\left[\left\langle\partial_{q^{\iota\kappa}}\mathcal{L}_{\lambda_1,\lambda_2}\right\rangle_{\mathcal{L}_{\lambda_1,\lambda_2}}\right] = -\frac{1}{2}\left[\lambda_2\right]^2 \mathbb{E}_z\left[\langle\xi^\iota\rangle_{\mathcal{L}_{\lambda_1,\lambda_2}}\langle\xi^\kappa\rangle_{\mathcal{L}_{\lambda_1,\lambda_2}}\right].$$

From this result, we find that the two remaining saddle-point equations are

$$\hat{q}^{\mu\nu} = \beta^2\alpha\left\langle\mathbb{E}_z\left[\langle\tau_\mu\rangle_{\mathcal{L}^O}\langle\tau_\nu\rangle_{\mathcal{L}^O}\right]\right\rangle_{\mathcal{M}_*}$$

$$q^{\mu\nu} = \mathbb{E}_{\xi^*}\mathbb{E}_z\left[\langle\xi^\mu\rangle_{\mathcal{L}^C}\langle\xi^\nu\rangle_{\mathcal{L}^C}\right].$$

**Appendix E. Saddle-point equations for Gaussian $\xi$**

In this Appendix, we take the prior $P(\xi)$ on the student pattern to be a standard Gaussian distribution. As per Section 3.1,

$$\mathcal{P}\left[\mathcal{L}^C\right](\xi;\xi^*,z) = \mathcal{Z}\left(\mathcal{L}^C\right)^{-1}\frac{1}{\sqrt{(2\pi)^P}}\exp\left(-\frac{1}{2}\sum_\mu\left[\xi^\mu\right]^2\right)\exp\left[\mathcal{L}^C(\xi;\xi^*,z)\right]$$

$$= \mathcal{Z}\left(\mathcal{L}^C\right)^{-1}\frac{1}{\sqrt{(2\pi)^P}}\exp\left(\mathcal{L}^C(\xi;\xi^*,z) - \frac{1}{2}\xi^T\xi\right),$$

where the second line is written in matrix notation. This expression will allow us to simplify the thermal averages $\langle\cdot\rangle_{\mathcal{L}^C}$ in the saddle-point equations (Eqs. (14)). By completing the square, $\mathcal{L}^C(\xi;\xi^*,z) - \frac{1}{2}\xi^T\xi$ can be written as

$$\mathcal{L}^C(\xi,\xi^*,z) - \frac{1}{2}\xi^T\xi = -\frac{1}{2}\left(\xi - [I+\hat{q}-\hat{s}]^{-1}\left[\hat{m}^T\xi^* + \text{diag}\left\{A(\hat{q})\frac{z+z^T}{2}\right\}\right]\right)^T$$

$$(I+\hat{q}-\hat{s})\left(\xi - [I+\hat{q}-\hat{s}]^{-1}\left[\hat{m}^T\xi^* + \text{diag}\left\{A(\hat{q})\frac{z+z^T}{2}\right\}\right]\right)$$

$$+ \frac{1}{2}\left[\hat{m}^T\xi^* + \text{diag}\left\{A(\hat{q})\frac{z+z^T}{2}\right\}\right]^T$$

$$[I+\hat{q}-\hat{s}]^{-1}\left[\hat{m}^T\xi^* + \text{diag}\left\{A(\hat{q})\frac{z+z^T}{2}\right\}\right],$$

where the function $\text{diag}(\hat{q})$ returns the diagonal of $\hat{q}$. We read out the mean and variance as

$$\langle\xi\rangle_{\mathcal{L}^C} = [I+\hat{q}-\hat{s}]^{-1}\left[\hat{m}^T\xi^* + \text{diag}\left\{A(\hat{q})\frac{z+z^T}{2}\right\}\right],$$

$$\left\langle\xi\xi^T\right\rangle_{\mathcal{L}^C} - \langle\xi\rangle_{\mathcal{L}^C}\langle\xi\rangle_{\mathcal{L}^C}^T = [I+\hat{q}-\hat{s}]^{-1}.$$

By evaluating the thermal averages $\langle\cdot\rangle_{\mathcal{L}^C}$ of the saddle-point equations, we then obtain

$$m = \mathbb{E}_{\xi^*}\left[\xi^*\xi^{*T}\hat{m}[I+\hat{q}-\hat{s}]^{-1}\right]$$

$$q = \mathbb{E}_{\xi^*}\left[[I+\hat{q}-\hat{s}]^{-1}\hat{m}^T\xi^*\xi^{*T}\hat{m}[I+\hat{q}-\hat{s}]^{-1}\right]$$

$$+ \mathbb{E}_z\left[[I+\hat{q}-\hat{s}]^{-1}\text{diag}\left\{A(\hat{q})\frac{z+z^T}{2}\right\}\text{diag}\left\{A(\hat{q})\frac{z+z^T}{2}\right\}^T[I+\hat{q}-\hat{s}]^{-1}\right]$$

$$s = [I+\hat{q}-\hat{s}]^{-1} + \mathbb{E}_{\xi^*}\left[[I+\hat{q}-\hat{s}]^{-1}\hat{m}^T\xi^*\xi^{*T}\hat{m}[I+\hat{q}-\hat{s}]^{-1}\right]$$

$$+ \mathbb{E}_z\left[[I+\hat{q}-\hat{s}]^{-1}\text{diag}\left\{A(\hat{q})\frac{z+z^T}{2}\right\}\text{diag}\left\{A(\hat{q})\frac{z+z^T}{2}\right\}^T[I+\hat{q}-\hat{s}]^{-1}\right].$$

The expected value over $z$ simplifies to

$$\mathbb{E}_z\left[\text{diag}\left\{A(\hat{q})\frac{z+z^T}{2}\right\}\text{diag}\left\{A(\hat{q})\frac{z+z^T}{2}\right\}^T\right]_{\mu\nu}$$

$$= \mathbb{E}_z\left[\sum_\kappa A_{\mu\kappa}(\hat{q})\frac{z_{\kappa\mu}+z_{\mu\kappa}}{2}\sum_\iota A_{\nu\iota}(\hat{q})\frac{z_{\iota\nu}+z_{\nu\iota}}{2}\right]$$

$$= \frac{1}{4} \mathbb{E}_z \left[ \sum_{\kappa,\iota} A_{\mu\kappa}(\hat{q}) A_{\nu\iota}(\hat{q}) z_{\kappa\mu} z_{\iota\nu} \right]$$

$$+ \frac{1}{4} \mathbb{E}_z \left[ \sum_{\kappa,\iota} A_{\mu\kappa}(\hat{q}) A_{\nu\iota}(\hat{q}) z_{\kappa\mu} z_{\nu\iota} \right]$$

$$+ \frac{1}{4} \mathbb{E}_z \left[ \sum_{\kappa,\iota} A_{\mu\kappa}(\hat{q}) A_{\nu\iota}(\hat{q}) z_{\mu\kappa} z_{\iota\nu} \right]$$

$$+ \frac{1}{4} \mathbb{E}_z \left[ \sum_{\kappa,\iota} A_{\mu\kappa}(\hat{q}) A_{\nu\iota}(\hat{q}) z_{\mu\kappa} z_{\nu\iota} \right]$$

$$= \frac{1}{2} \delta_{\mu\nu} \sum_\iota \left[ A_{\mu\iota}(\hat{q}) \right]^2 + \frac{1}{2} \left[ A_{\mu\nu}(\hat{q}) \right]^2$$

$$= \hat{q}^{\mu\nu}.$$

In the end, we find

$$\begin{aligned}
m &= Q\hat{m} \left[ I + \hat{q} - \hat{s} \right]^{-1} \\
q &= \left[ I + \hat{q} - \hat{s} \right]^{-1} \hat{m}^T Q \hat{m} \left[ I + \hat{q} - \hat{s} \right]^{-1} \\
&\quad + \left[ I + \hat{q} - \hat{s} \right]^{-1} \hat{q} \left[ I + \hat{q} - \hat{s} \right]^{-1} \\
s &= \left[ I + \hat{q} - \hat{s} \right]^{-1} + \left[ I + \hat{q} - \hat{s} \right]^{-1} \hat{m}^T Q \hat{m} \left[ I + \hat{q} - \hat{s} \right]^{-1} \\
&\quad + \left[ I + \hat{q} - \hat{s} \right]^{-1} \hat{q} \left[ I + \hat{q} - \hat{s} \right]^{-1} .
\end{aligned} \quad \text{(E.1)}$$

**Appendix F. Critical load**

In the paramagnetic phase, the order parameters all vanish. The paramagnetic to ferromagnetic phase transition of the student RBM is the line where the paramagnetic solution of the saddle-point equations (see Eq. (14)) becomes unstable to leading order in the order parameters (see Fig. 10). As a consequence of Eq. (D.1), any Hamiltonian of the form $\mathcal{H}(\xi) = \frac{1}{2} \sum_{\mu \neq \nu} J_{\mu\nu} \xi^\mu \xi^\nu + \sum h_\mu \xi^\mu$ has

$$\langle \xi^\mu \rangle_\mathcal{H} \approx h_\mu$$

to first order in the parameters $J_{\mu\nu}$ and $h_\mu$. Therefore, given that the prior on $\xi$ has a mean of zero, we have

$$\mathbb{E}_z \left[ \xi^{*\gamma} \langle \xi^\mu \rangle_{\mathcal{L}_{\lambda_1,\lambda_2}} \right] \approx \lambda_1 \lambda_2 \mathbb{E}_z \left[ \xi^{*\gamma} \sum_\rho m^{\rho\mu} \xi^{*\rho} \right]$$

$$= \lambda_1 \lambda_2 \sum_\rho m^{\rho\mu} \xi^{*\gamma} \xi^{*\rho}$$

to first order in $m$, $s$ and $q$. The saddle-point equations for $\hat{m}$ and $m$ then simplify to

$$\begin{aligned}
\hat{m}^{\gamma\mu} &= \beta^* \beta \alpha \left\langle \mathbb{E}_z \left[ \tau_{*\gamma} \langle \tau_\mu \rangle_{\mathcal{L}^O} \right] \right\rangle_{\mathcal{M}_*} \\
&= \left[ \beta^* \beta \right]^2 \alpha \sum_\rho \langle \tau_{*\gamma} \tau_{*\rho} \rangle_{\mathcal{M}_*} m^{\rho\mu} \\
&= \left[ \beta^* \beta \right]^2 \alpha \sum_\rho \mathcal{R}_{\mu\rho} m^{\rho\mu} \\
m^{\gamma\mu} &= \mathbb{E}_{\xi^*} \mathbb{E}_z \left[ \xi^{*\gamma} \langle \xi^\mu \rangle_{\mathcal{L}^C} \right] \\
&= \sum_\rho \mathbb{E}_{\xi^*} \left[ \xi^{*\gamma} \xi^{*\rho} \right] \hat{m}^{\rho\mu} \\
&= \sum_\rho \mathcal{Q}_{\mu\rho} \hat{m}^{\rho\mu},
\end{aligned}$$

where $\mathcal{Q}$ and $\mathcal{R}$ are the covariance matrices of $\xi^*$ and $\mathcal{M}_*$, respectively. We see that the behavior of $\hat{m}$ and $m$ is much simpler near criticality than at an arbitrary location in the phase diagram. In fact, the stationary values of $\hat{m}$ and $m$ do not depend on the other order parameters. We can even rewrite the equations for $\hat{m}$ and $m$ in the more compact form

$$m^{\gamma\mu} = \left[ \beta^* \beta \right]^2 \alpha \sum_{\iota,\kappa} \mathcal{Q}_{\mu\iota} \mathcal{R}_{\iota\kappa} m^{\kappa\mu}.$$

Let the largest eigenvalue of $S_{\mu\kappa} = \sum_\iota \mathcal{Q}_{\mu\iota} \mathcal{R}_{\iota\kappa}$ be $\lambda^S_{\max}$. As known from stability theory (Strogatz, 2018), the paramagnetic solution $m^{\gamma\mu} = 0$ is unstable when $[\beta^* \beta]^2 \alpha \, S$ has at least one eigenvalue larger than 1. In other words, the student is able to learn the teacher patterns above a critical load of

$$\alpha_{\text{crit}} = \frac{1}{[\beta^* \beta]^2 \lambda^S_{\max}}.$$

**Appendix G. Saddle-point equations in the absence of correlations**

In the absence of correlations, we have $\mathcal{Q} = \mathbf{I}$. The effective Hamiltonian $\mathcal{M}_*$ (see Eq. (7)) then simplifies to

$$\mathcal{M}_*(\tau_*) = \frac{1}{2} P^* \left[ \beta^* \right]^2,$$

so the expectation $\langle \cdot \rangle_{\mathcal{M}_*}$ is uniform. When $P \leq P^*$, i.e. the student has at most the same number of hidden units as the teacher, we make the ansatz

$$\begin{aligned}
m^{\gamma\mu} &= \delta_{\gamma\mu} m, & \hat{m}^{\gamma\mu} &= \delta_{\gamma\mu} \hat{m}, \\
s^{\mu\nu} &= \delta_{\mu\nu}, & \hat{s}^{\mu\nu} &= 0, \\
q^{\mu\nu} &= \delta_{\mu\nu} q, & \hat{q}^{\mu\nu} &= \delta_{\mu\nu} \hat{q},
\end{aligned} \quad \text{(G.1)}$$

under which $\mathcal{M}$ (Eq. (8)) and $\mathcal{L}_{\lambda_1,\lambda_2}$ (Eq. (11)) respectively simplify to

$$\mathcal{M}(\tau) = \frac{1}{2} P \beta^2 \quad \text{and}$$

$$\mathcal{L}_{\lambda_1,\lambda_2}(\xi^*, \xi, z; m, s, q) = \frac{1}{2} P \left[ \lambda_2 \right]^2 (1 - q) + \sum_{\mu=1}^P \left( \lambda_1 \lambda_2 \, m \xi^{*\mu} + \lambda_2 \sqrt{q} z_\mu \right) \xi^\mu.$$

The spins $\xi^\mu$ do not interact with one another in $\mathcal{L}_{\lambda_1,\lambda_2}(\xi^*, \xi, z; m, s, q)$. In other words, they are independent with respect to the Gibbs distribution with Hamiltonian $\mathcal{L}_{\lambda_1,\lambda_2}$. Therefore, all of the spins with $\nu \neq \mu$ can be marginalized from the thermal average $\langle \xi^\mu \rangle_{\mathcal{L}_{\lambda_1,\lambda_2}}$. In particular, when the student patterns $\xi$ are binary random variables, we obtain

$$\begin{aligned}
\langle \xi^\mu \rangle_{\mathcal{L}_{\lambda_1,\lambda_2}} &= \sum_{\xi^\mu = \pm 1} \frac{\exp\left( \left[ \lambda_1 \lambda_2 \, m \xi^{*\mu} + \lambda_2 \sqrt{q} z_\mu \right] \xi^\mu \right) \xi^\mu}{\exp\left( \lambda_1 \lambda_2 \, m \xi^{*\mu} + \lambda_2 \sqrt{q} z_\mu \right) + \exp\left( -\lambda_1 \lambda_2 \, m \xi^{*\mu} - \lambda_2 \sqrt{q} z_\mu \right)} \\
&= \tanh\left( \lambda_1 \lambda_2 \, m \xi^{*\mu} + \lambda_2 \sqrt{q} z_\mu \right).
\end{aligned}$$

Similarly, the hidden unit spins $\tau_\mu$ are independent with respect to the Gibbs distribution with Hamiltonian $\mathcal{M}(\tau)$. By the independence of all these spins, the saddle-point equations (Eqs. (14)) with binary $\xi$ reduce to

$$\begin{aligned}
\hat{m}^{\gamma\mu} &= \beta^* \beta \alpha \, \delta_{\gamma\mu} \left\langle \mathbb{E}_z \left[ \tau_{*\mu} \tanh\left( \beta^* \beta m \tau_{*\mu} + \beta \sqrt{q} z_\mu \right) \right] \right\rangle_{\mathcal{M}_*} \\
\hat{s}^{\mu\nu} &= 0 \\
\hat{q}^{\mu\nu} &= \beta^2 \alpha \, \delta_{\mu\nu} \left\langle \mathbb{E}_z \left[ \tanh^2\left( \beta^2 \, m \tau_{*\mu} + \beta \sqrt{q} z_\mu \right) \right] \right\rangle_{\mathcal{M}_*} \\
m^{\gamma\mu} &= \delta_{\gamma\mu} \, \mathbb{E}_{\xi^*} \mathbb{E}_z \left[ \xi^{*\mu} \tanh\left( \hat{m} \xi^{*\mu} + \sqrt{\hat{q}} z_\mu \right) \right] \\
s^{\mu\nu} &= 0 \\
q^{\mu\nu} &= \delta_{\mu\nu} \, \mathbb{E}_{\xi^*} \mathbb{E}_z \left[ \tanh^2\left( \hat{m} \xi^{*\mu} + \sqrt{\hat{q}} z_\mu \right) \right].
\end{aligned}$$

Assume the teacher patterns $\xi^*$ are also binary. Since $\tanh$ is an odd function, we factor the spins out of it according to $\tanh(\xi^{*\mu} y) = \xi^{*\mu} \tanh(y)$ and use the change of variables $z = \xi^{*\mu} z_\mu$ to simplify the saddle-point equations to

$$\begin{aligned}
\hat{m} &= \beta^* \beta \alpha \, \mathbb{E}_z \left[ \tanh\left( \beta^* \beta m + \beta \sqrt{q} z \right) \right] \\
\hat{q} &= \beta^2 \alpha \, \mathbb{E}_z \left[ \tanh^2\left( \beta^2 \, m + \beta \sqrt{q} z \right) \right] \\
m &= \mathbb{E}_z \left[ \tanh\left( \hat{m} + \sqrt{\hat{q}} z \right) \right] \\
q &= \mathbb{E}_z \left[ \tanh^2\left( \hat{m} + \sqrt{\hat{q}} z \right) \right].
\end{aligned}$$

On the Nishimori line $\beta^* = \beta$, we have

$$\hat{m} = \beta^2 \alpha \, \mathbb{E}_z \left[ \tanh\left( \beta^2 \, m + \beta \sqrt{m} z \right) \right]$$

$$m = \mathbb{E}_z \left[ \tanh\left( \hat{m} + \sqrt{\hat{m}} z \right) \right].$$

When $P > P^*$, the saddle-point equations are slightly different. We make the ansatz

$$m^{\gamma\mu} = \delta_{\gamma\mu} m, \qquad \hat{m}^{\gamma\mu} = \delta_{\gamma\mu}\hat{m},$$
$$s^{\mu\nu} = \delta_{\mu\nu}, \qquad \hat{s}^{\mu\nu} = 0,$$
$$q^{\mu\nu} = \begin{cases} \delta_{\mu\nu} q & \mu,\nu \le P^* \\ \delta_{\mu\nu} g & \text{otherwise,} \end{cases} \qquad \hat{q}^{\mu\nu} = \begin{cases} \delta_{\mu\nu}\hat{q} & \mu,\nu \le P^* \\ \delta_{\mu\nu}\hat{g} & \text{otherwise,} \end{cases} \quad \text{(G.2)}$$

and obtain

$$\mathcal{L}_{\lambda_1,\lambda_2}\left(\xi^*,\xi,z;m,s,q\right) = \sum_{\mu=1}^{P^*} \left(\lambda_1\lambda_2\, m\xi^{*\mu} + \lambda_2\sqrt{q}z_\mu\right)\xi^\mu + \sum_{\mu=P^*+1}^{P} \lambda_2\sqrt{g}z_\mu\xi^\mu.$$

Following the same steps as for $P \le P^*$, we get

$$\hat{m} = \beta^*\beta\alpha\, \mathbb{E}_z\left[\tanh\left(\beta^*\beta m + \beta\sqrt{q}z\right)\right]$$
$$\hat{q} = \beta^2\alpha\, \mathbb{E}_z\left[\tanh^2\left(\beta^2\, m + \beta\sqrt{q}z\right)\right]$$
$$\hat{g} = \beta^2\alpha\, \mathbb{E}_z\left[\tanh^2\left(\beta\sqrt{g}z\right)\right]$$
$$m = \mathbb{E}_z\left[\tanh\left(\hat{m} + \sqrt{\hat{q}}z\right)\right]$$
$$q = \mathbb{E}_z\left[\tanh^2\left(\hat{m} + \sqrt{\hat{q}}z\right)\right]$$
$$g = \mathbb{E}_z\left[\tanh^2\left(\sqrt{\hat{g}}z\right)\right].$$

When $\beta = \beta^*$, these equations reduce to

$$\hat{m} = \beta^2\alpha\, \mathbb{E}_z\left[\tanh\left(\beta^2\, m + \beta\sqrt{m}z\right)\right]$$
$$\hat{g} = \beta^2\alpha\, \mathbb{E}_z\left[\tanh^2\left(\beta\sqrt{g}z\right)\right]$$
$$m = \mathbb{E}_z\left[\tanh\left(\hat{m} + \sqrt{\hat{m}}z\right)\right]$$
$$g = \mathbb{E}_z\left[\tanh^2\left(\sqrt{\hat{g}}z\right)\right].$$

When the student patterns $\xi$ are Gaussian random variables, we find instead

$$\hat{m} = \beta^*\beta\alpha\, \mathbb{E}_z\left[\tanh\left(\beta^*\beta m + \beta\sqrt{q}z\right)\right]$$
$$\hat{q} = \beta^2\alpha\, \mathbb{E}_z\left[\tanh^2\left(\beta^2\, m + \beta\sqrt{q}z\right)\right]$$
$$\hat{g} = \beta^2\alpha\, \mathbb{E}_z\left[\tanh^2\left(\beta\sqrt{g}z\right)\right]$$
$$m = \frac{\hat{m}}{1+\hat{q}}$$
$$q = \frac{\hat{m}^2}{(1+\hat{q})^2} + \frac{\hat{q}}{(1+\hat{q})^2}$$
$$g = \frac{\hat{g}}{(1+\hat{g})^2}.$$

When $\beta = \beta^*$, these equations reduce to

$$\hat{m} = \beta^2\alpha\, \mathbb{E}_z\left[\tanh\left(\beta^2\, m + \beta\sqrt{m}z\right)\right]$$
$$\hat{g} = \beta^2\alpha\, \mathbb{E}_z\left[\tanh^2\left(\beta\sqrt{g}z\right)\right]$$
$$m = \frac{\hat{m}}{1+\hat{m}}$$
$$g = \frac{\hat{g}}{(1+\hat{g})^2}.$$

### Appendix H. Effect of uniform correlations

We introduce uniform correlations in the teacher patterns by fixing the covariance matrix $\mathcal{Q}$ of $\xi^*$ to $\mathcal{Q}_{\gamma\rho} = \delta_{\gamma\rho} + (1-\delta_{\gamma\rho})c$, where $c \in [0,1)$. Given this particular $\mathcal{Q}$, the Hamiltonian $\mathcal{M}_*$ (see Eq. (7)) simplifies to

$$\mathcal{M}_*(\tau_*) = \frac{1}{2}[\beta^*]^2 c \sum_{\gamma \ne \rho} \tau_{*\gamma}\tau_{*\rho} + \frac{1}{2} P^* [\beta^*]^2. \quad \text{(H.1)}$$

The interaction between any two spins $\tau_{*\gamma}$ and $\tau_{*\rho}$ does not depend on the sites $\gamma \ne \rho$. Therefore, the covariance matrix $\mathcal{R}$ of $\tau_*$ has the same form as $\mathcal{Q}$, but with a different coefficient $d$ outside the diagonal. $\mathcal{S} = \mathcal{Q}\mathcal{R}$ then reduces to

$$\mathcal{S}_{\gamma\rho} = \sum_{\tau} \left(c + (1-c)\delta_{\gamma\tau}\right)\left(d + (1-d)\delta_{\tau\rho}\right)$$

$$= P^* cd + c(1-d) + (1-c)d + (1-c)(1-d)\delta_{\gamma\rho}$$

Any $P^* \times P^*$ matrix of the form $\mathcal{A}_{\gamma\rho} = a + b\delta_{\gamma\rho}$ has eigenvalues

$$\lambda_1^{\mathcal{A}} = P^* a + b \text{ with corresponding eigenvector } e = \frac{1}{\sqrt{P^*}}\begin{bmatrix}1 & \ldots & 1\end{bmatrix},$$

$$\lambda_2^{\mathcal{A}} = b \text{ with corresponding eigenspace } \left\{x \in \mathbb{R}^{P^*} \,\Big|\, \sum_i x_i = 0\right\}.$$

Therefore, $\mathcal{S}$ has eigenvalues

$$\lambda_1^{\mathcal{S}} = P^*\left(P^* cd + c(1-d) + (1-c)d\right) + (1-c)(1-d),$$
$$\lambda_2^{\mathcal{S}} = (1-c)(1-d).$$

$d$ is positive because the interaction between any two spins $\tau_{*\gamma}$ and $\tau_{*\rho}$ is positive (see Eq. (H.1) and Griffiths (1967)). Therefore, the largest eigenvalue $\lambda_{\max}^{\mathcal{S}}$ of $\mathcal{S}$ is $\lambda_1^{\mathcal{S}}$. In sum,

$$\lambda_{\max}^{\mathcal{S}} = \lambda_1^{\mathcal{S}} = (P^*-1)^2 cd + (P^*-1)(c+d) + 1.$$

### Appendix I. Numerical methods

We solve the saddle-point equations (Eqs. (14)) by numerical iteration. To be more specific, we iterate

$$\hat{m}^{\gamma\mu}(t+1) = \hat{m}^{\gamma\mu}(t) + \Delta t\left(\beta^*\beta\alpha\, \mathbb{E}_{\mathcal{M}_*}\mathbb{E}_z\left[\tau_{*\gamma}\langle\tau_\mu\rangle_{\mathcal{L}^O(t+1)}\right] - \hat{m}^{\gamma\mu}(t)\right)$$
$$\hat{s}^{\mu\nu}(t+1) = \hat{s}^{\mu\nu}(t) + \Delta t\left(\beta^2\alpha\left(\mathbb{E}_{\mathcal{M}_*}\mathbb{E}_z\left[\langle\tau_\mu\tau_\nu\rangle_{\mathcal{L}^O(t)}\right] - \langle\tau_\mu\tau_\nu\rangle_{\mathcal{M}(t)}\right) - \hat{s}^{\mu\nu}(t)\right)$$
$$\hat{q}^{\mu\nu}(t+1) = \hat{q}^{\mu\nu}(t) + \Delta t\left(\beta^2\alpha\mathbb{E}_{\mathcal{M}_*}\mathbb{E}_z\left[\langle\tau_\mu\rangle_{\mathcal{L}^O(t)}\langle\tau_\nu\rangle_{\mathcal{L}^O(t)}\right] - \hat{q}^{\mu\nu}(t)\right)$$
$$m^{\gamma\mu}(t+1) = m^{\gamma\mu}(t) + \Delta\tau\left(\mathbb{E}_{\xi^*}\mathbb{E}_z\left[\xi^{*\gamma}\langle\xi^\mu\rangle_{\mathcal{L}^C(t+1)}\right] - m^{\gamma\mu}(t)\right)$$
$$s^{\mu\nu}(t+1) = s^{\mu\nu}(t) + \Delta\tau\left(\mathbb{E}_{\xi^*}\mathbb{E}_z\left[\langle\xi^\mu\xi^\nu\rangle_{\mathcal{L}^C(t+1)}\right] - s^{\mu\nu}(t)\right)$$
$$q^{\mu\nu}(t+1) = q^{\mu\nu}(t) + \Delta\tau\left(\mathbb{E}_{\xi^*}\mathbb{E}_z\left[\langle\xi^\mu\rangle_{\mathcal{L}^C(t+1)}\langle\xi^\nu\rangle_{\mathcal{L}^C(t+1)}\right] - q^{\mu\nu}(t)\right),$$

with time steps $\Delta t, \Delta\tau \in (0,1]$. By construction, Eqs. (14) are a fixed point of the iteration. Empirically, the iteration is the most stable when one of the two time steps is equal to one and the other one is small. Even then, it is still occasionally unstable at large $\alpha$, large teacher pattern correlations and low $T$, which introduces a few easily identifiable spurious discontinuities in Figs. 10. We symmetrize the initial conditions of $q$ before the iteration because any $q$ solving the saddle-point equations must be symmetric (see Eqs. (14)).

We use Monte Carlo integration over whitened samples to estimate the Gaussian expectations $\mathbb{E}_z[\cdot]$. We enforce the sample means to be 0 by symmetrizing around the origin each set of samples $z_{\mu\nu}$ that approximates the corresponding integral over $z_{\mu\nu}$. We then constrain the sample variances to 1 using Cholesky whitening (Agnan Kessy & Strimmer, 2018).

$\mathcal{L}_{\lambda_1,\lambda_2}(\xi,\xi^*,z)$ is a complex-valued function because it involves the square root of a real number that is not necessarily positive. Therefore, the thermal averages $\langle\cdot\rangle_{\mathcal{L}^C}$ and $\langle\cdot\rangle_{\mathcal{L}^O}$, are also complex-valued. By the symmetry of the Gaussian variables in the saddle point equations (see Eqs. (14)), the imaginary part of each thermal average is equal to $a$ and $-a$ with the same probability. Therefore, the exact Gaussian expectations must be real-valued. However, in practice, standard Monte Carlo integration for evaluating $\mathbb{E}_z[g(z)]$ with a finite number of samples is very unlikely to randomly produce both $g(z)$ and its complex conjugate $\bar{g}(z)$. Therefore, we replace $\mathbb{E}_z[\cdot]$ by $\mathbb{E}_z[\text{Re}(\cdot)]$ when solving the saddle-point equations numerically. This adjustment significantly increases the stability of the numerical iteration.

### Appendix J. Supplementary figures

This Appendix contains a graph of the free entropy difference of the PSB and partial PSB solutions of Eqs. (14), as well as some plots of the Mattis magnetization $m$ and spin-glass overlap $q$ obtained for real-valued student patterns with a standard Gaussian prior. The former supports some claims made in Section 3.2.1, but is not strictly necessary to understand the paper. The latter are not shown in the main text because they look similar to the $m$ and $q$ obtained for binary student patterns with a uniform binary prior. We simplified the saddle-point equations (Eqs. (14)) according to Appendix E in order to make them.

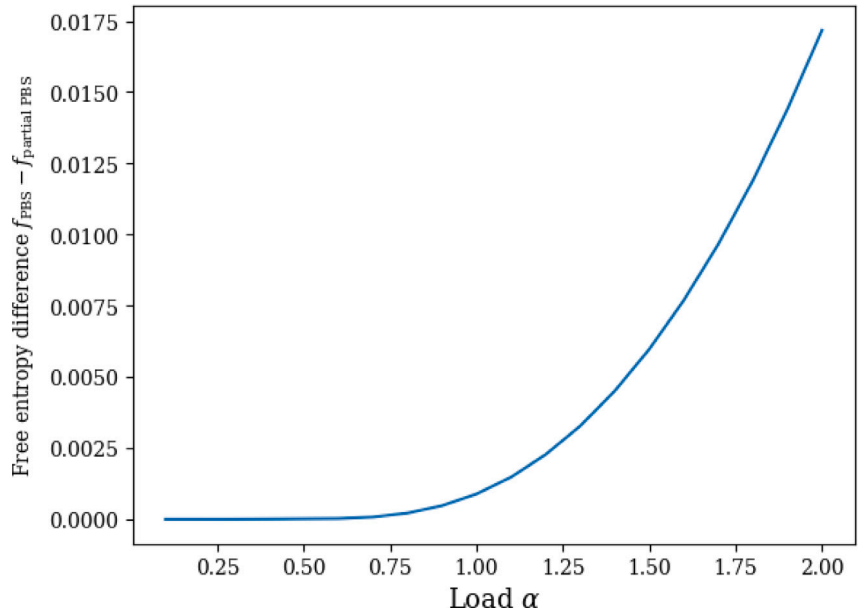

**Fig. J.17.** Free entropy difference of the so-called PSB and partial PSB solutions of Eqs. (14) shown in Figs. 2 and 3. This plot was made using $P^* = 2$ and $P = 3$, but the free entropy of $P^* = 3$ and $P = 4$ looks identical.

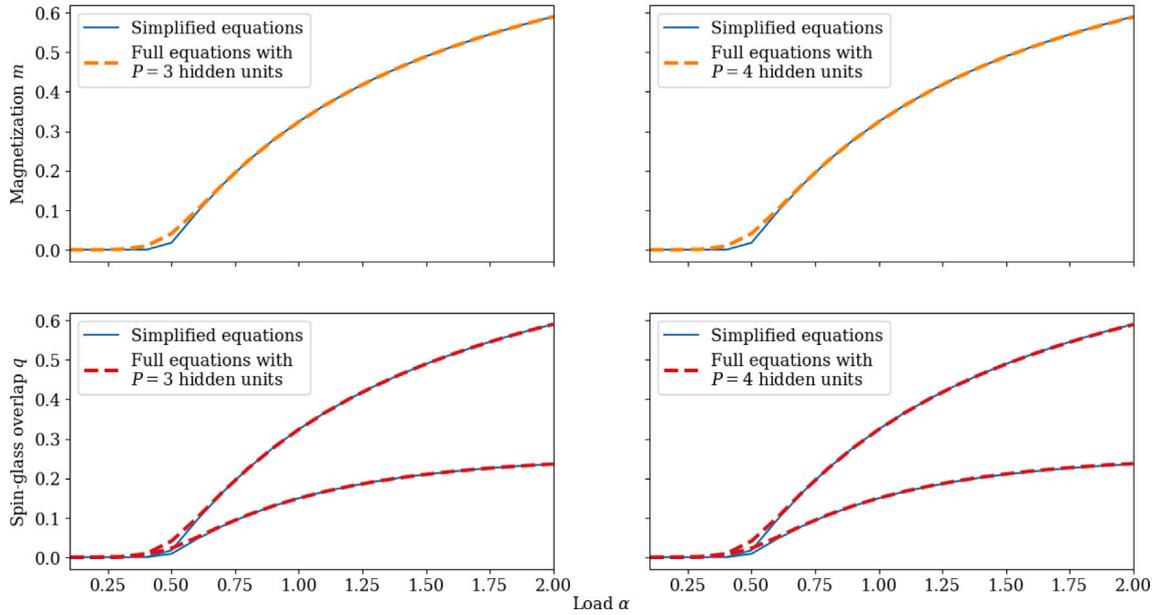

**Fig. J.18.** Permutation symmetry breaking (PSB) solution of Eqs. (14) for real-valued student patterns with a standard Gaussian prior and teacher pattern covariance $Q = I$, in red and orange, compared against the solution of Eqs. (23), in blue. We plot the Mattis magnetization $m$ in the top row, and the SG overlap $q$ in the bottom row. The magnetization plots and the top lines of the SG overlap plots show that the student patterns that converge to teacher patterns have the same $m$ and $q$ as the solution of Eqs. (23), and thus also satisfy $m = q$. Conversely, the bottom lines of the SG overlap plots student patterns that do not converge to a teacher pattern have a spin-glass overlap of $g$ as in Eqs. (23). The top branch of $q$ is We use $P = 3$ and $P^* = 2$ in the left column and $P = 4$ and $P^* = 3$ in the right column. All plots have $\beta^* = \beta = 1.2$.

**Data availability**

The code and hyperparameter values of the training algorithms used to make the figures are available at the following public Github repository (Thériault, 2025).

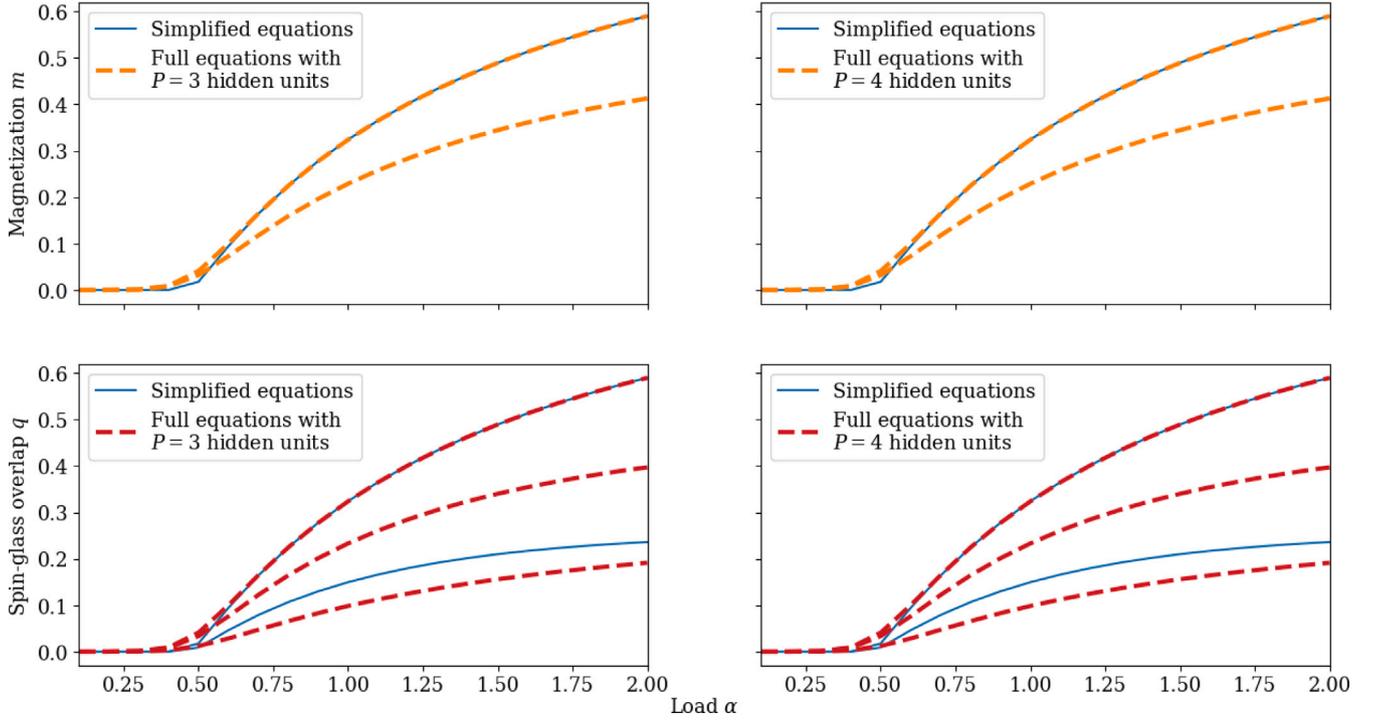

**Fig. J.19.** Partial permutation symmetry breaking (partial PSB) solutions of Eqs. (14) for real-valued student patterns with a standard Gaussian prior and teacher pattern covariance $\mathcal{Q} = \mathbf{I}$, in red and orange, compared against the solution of Eqs. (23), in blue. We plot the Mattis magnetization $m$ in the top row, and the SG overlap $g$ in the bottom row. The top lines of the plots show that the student patterns $\xi^\mu_{\text{PSB}}$ that converge to teacher patterns one-to-one have the same $m$ and $q$ as the solution of Eqs. (23), and thus also satisfy $m = q$. Conversely, the other lines show that the student patterns $\xi^\mu_{\text{PS}}$ that converge to a common teacher pattern have a smaller $m$ and a different $q$. To be more precise, the central and bottom branches of $q$ are the spin-glass order parameters corresponding to $Q(\xi^{1\mu}_{PS}, \xi^{2\mu}_{PS})$ and $Q(\xi^{1\mu}_{PS}, \xi^{2\nu}_{PS})$ with $\mu \neq \nu$, respectively (see Section 2). They are both different from the $g$ of Eq. (23). The Mattis magnetization and SG overlaps omitted from this Figure all vanish. We use $P = 3$ and $P^* = 2$ in the left column and $P = 4$ and $P^* = 3$ in the right column. All plots have $\beta^* = \beta = 1.2$.

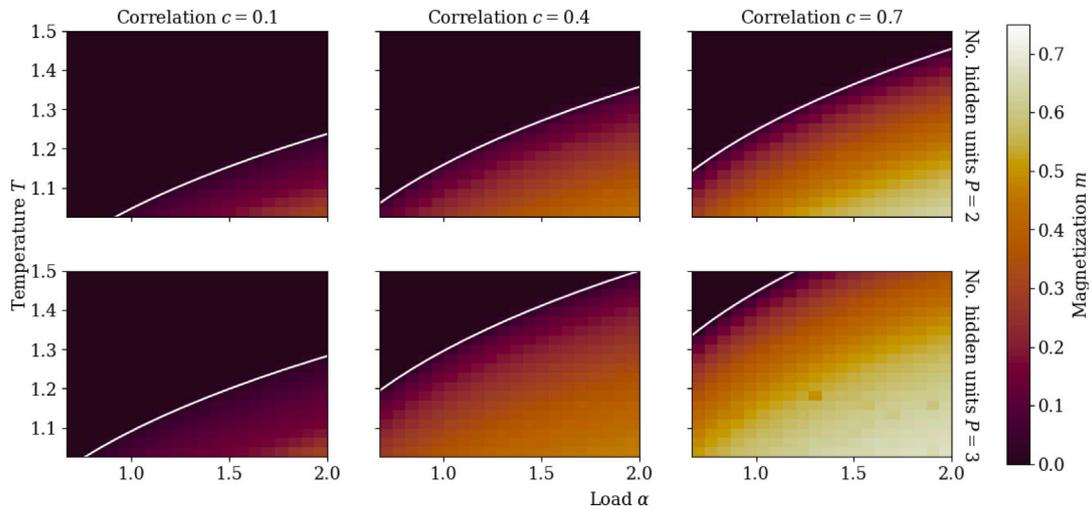

**Fig. J.20.** Mattis magnetization $m$ for $\beta = \beta^*$ and $P = P^*$ as a function of the number of hidden units $P$, the correlation $c$, the temperature $T$ and the data load $\alpha$. $m$ is obtained by solving Eqs. (14) numerically for real-valued student patterns with a standard Gaussian prior and teacher pattern covariance $\mathcal{Q}_{\mu\nu} = \delta_{\mu\nu} + (1 - \delta_{\mu\nu})c$, where $c \in [0, 1)$ (see Appendix H). The top and bottom rows feature $P = 2$ and $P = 3$, respectively. The white lines mark the phase transition of Eq. (18) with $\lambda^S_{\max}$ given by Eq. (24). The speckles in the plots with $P = 3$, $c = 0.4$ and $P = 3$, $c = 0.7$ are due to the saddle-point iteration converging to a different solution of Eqs. E than at neighboring $\alpha$ and $T$.

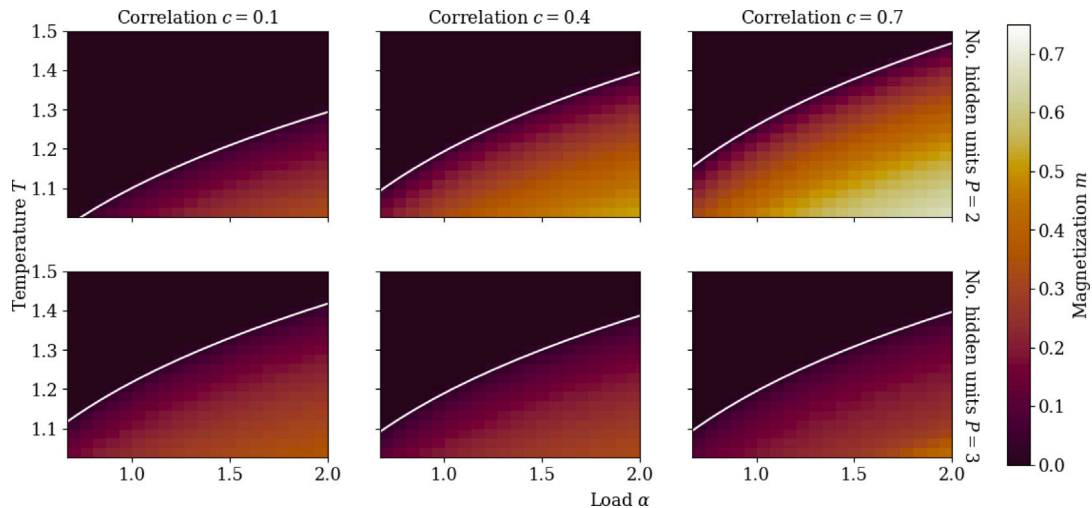

**Fig. J.21.** Mattis magnetization $m$ for $\beta = \beta^*$ and $P = P^*$ as a function of the number of hidden units $P$, the correlation $c$, the temperature $T$ and the data load $\alpha$. $m$ is obtained by solving Eqs. (14) for real-valued student patterns with a standard Gaussian prior and teacher pattern covariance $\mathcal{Q}_{\mu\nu} \sim \mathcal{W}(c, P)$, where $c \in [0, 1)$ (see Appendix A.2). The top and bottom rows feature $P = 2$ and $P = 3$, respectively. The white lines mark the phase transition of Eq. (18) with $\lambda^S_{\max}$ given by Eq. (24).